\documentclass[runningheads]{llncs}
\usepackage{graphicx}
\usepackage{amsmath}
\usepackage{colortbl}
\usepackage{color,soul}
\usepackage{booktabs,caption}
\usepackage{cite}
\usepackage[ruled,vlined,linesnumbered]{algorithm2e}
\usepackage{caption}
\usepackage{subcaption}
\usepackage{amssymb}
\usepackage{bbm}
\usepackage{mathtools}
\usepackage[round,authoryear]{natbib}
\usepackage{xcolor}
\usepackage{hyperref}
\hypersetup{
    colorlinks,
    linkcolor={red!50!black},
    citecolor={blue!50!black},
    urlcolor={blue!80!black}
}
\begin{document}
\renewcommand{\labelitemi}{$\bullet$}

\title{Explaining and visualizing black-box models through counterfactual paths}

\titlerunning{Explaining and visualizing black-box models through counterfactual paths}
%
\author{Bastian Pfeifer \inst{1}
\and 
Mateusz Krzyzinski \inst{2} \and
Hubert Baniecki \inst{3} \and
Anna Saranti   \inst{1,4} \and \\
Andreas Holzinger  \inst{1,4} \and
Przemyslaw Biecek \inst{2,3} \\
}
\authorrunning{Pfeifer et. al}

%

%

\institute{Institute for Medical Informatics, Statistics and Documentation. \\ Medical University Graz, Austria \\ \texttt{bastian.pfeifer@medunigraz.at}\\
\and
MI2.AI, Warsaw University of Technology, Poland
\and
MI2.AI, University of Warsaw, Poland
\and
Human-Centered AI Lab, University of Natural Resources and Life Sciences, Vienna, Austria}

%
\maketitle              
\begin{abstract}

Explainable AI (XAI) is an increasingly important area of machine learning research, which aims to make black-box models transparent and interpretable. In this paper, we propose a novel approach to XAI that uses the so-called \emph{counterfactual paths} generated by conditional permutations of features. The algorithm measures feature importance by identifying sequential permutations of features that most influence changes in model predictions. It is particularly suitable for generating explanations based on counterfactual paths in knowledge graphs incorporating domain knowledge. \emph{Counterfactual paths} introduce an additional graph dimension to current XAI methods in both explaining and visualizing black-box models. Experiments with synthetic and medical data demonstrate the practical applicability of our approach.


\keywords{explainable machine learning, knowledge graph, feature importance, counterfactual explanation}
\end{abstract}
\section{Introduction}

\begin{figure}
     \centering
         \includegraphics[width=\textwidth]{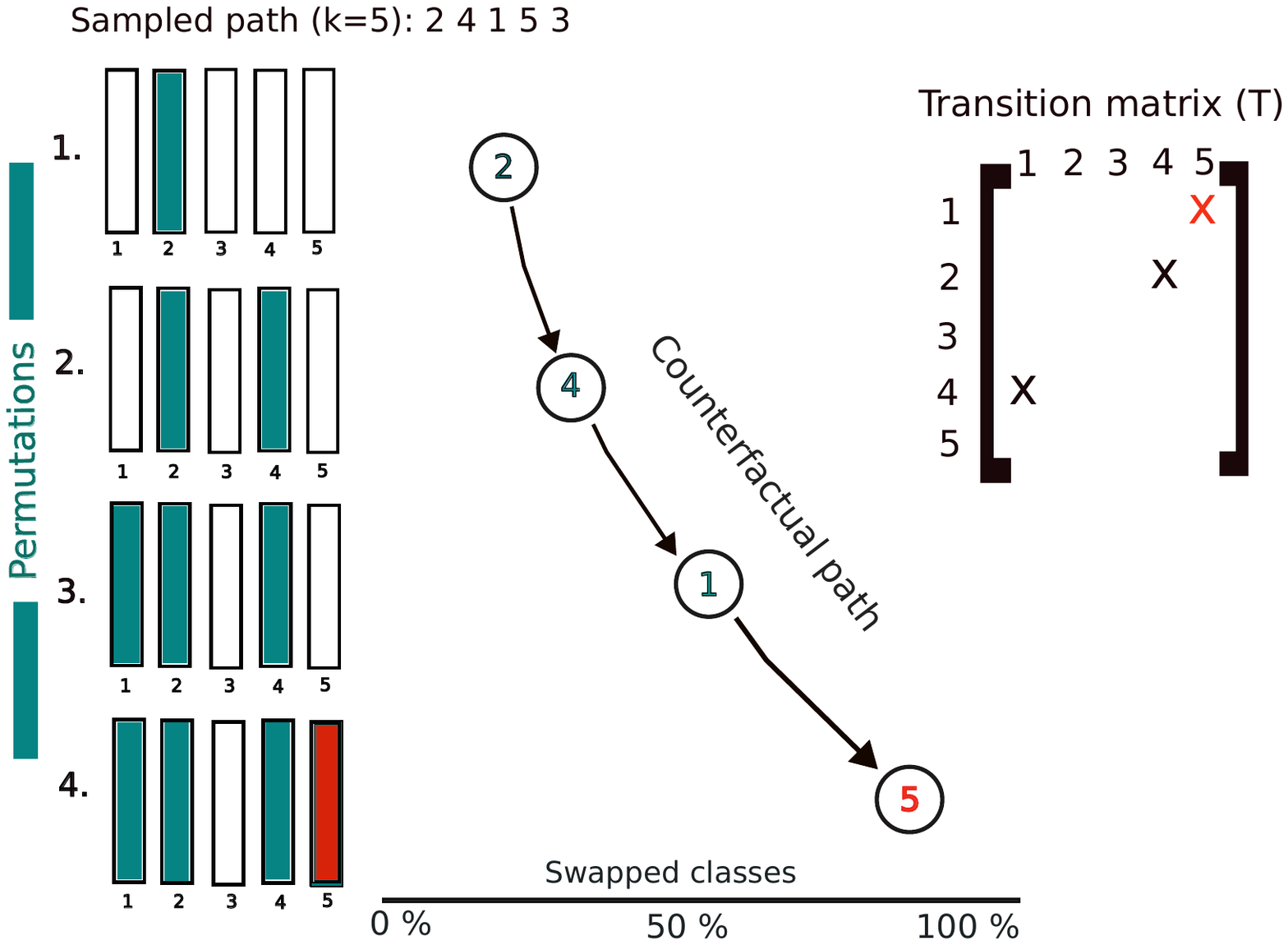}
         \caption{Counterfactual path. Shown is the core principle of our proposed algorithm. The input data is in tabular form. A certain $k$ number of features is sampled, forming a candidate counterfactual path. These features are then subsequently permuted according to the ordering reflected by the candidate path. The swapped classes are tracked, and in case they exceed a certain threshold, the permutation procedure terminates and the detected counterfactual path is stored. According to the edges of that path a transition matrix ($\mathbf{T}$) is formed, from which feature importances are derived.}
        \label{fig:abstract}
\end{figure}


Explainable AI (XAI) methods are becoming a promising solution to the emerging challenge of providing human-understandable justifications for AI decisions \citep{holzinger2022xai}, e.g. in medicine \citep{GOZZI2022108053,XU2022109200,KRZYZINSKI2023110234}, biology \citep{ruiz_explainable_2020} or finance \citep{Szepannek2022}. One of the key problems in model explanation is determining the importance of variables, and to this end, many methods have been proposed either specific to some model families or model agnostic, here we focus on the latter. One approach to XAI is through counterfactual explanations, which involve generating alternative scenarios that could have led to different prediction outcomes. Counterfactual explanations help users understand how an AI model arrived at a particular decision and what factors influenced that decision \citep{chou2022counterfactuals}. Another widely adopted approach is feature importance methods that express the reliance of model predictive performance on a specific feature in data \citep{fisher2019all,casaliccio2019visualizing}. 

However, these methods most often present the importance for variables separately, without taking into account, often complex relationships between variables such as interactions or correlations. To challenge this status quo, we introduce the counterfactual paths (CPATH) algorithm for model-agnostic global explanations of machine learning predictive models trained on tabular data. This algorithm is inspired by both counterfactual explanations and permutational feature importance (see Figure \ref{fig:abstract} for a graphical illustration).
Compared to classical feature importance methods, CPATH provides additional graph information about the counterfactual dependence of the black-box model on particular features, and thus can help to uncover its decision-making process.

Counterfactual paths represent the relationships between input features and the model's output, allowing users to explore how changing individual features or their combinations affect the model's predictions. One of the main advantages of counterfactual paths is their ability to provide insights into how the model works and to identify potential biases or confounding factors that may affect its predictions. These insights can then improve the model's accuracy and robustness.

Furthermore, counterfactual paths and their visualization could provide a more intuitive and interpretable explanation of the model's behaviour than traditional feature importance methods. For example, rather than simply providing a list of the top features that contribute to the model's predictions, counterfactual paths can show how changes to specific combinations of features lead to changes in the output of the model. This can help users to better understand the underlying patterns in the data and the overall behaviour of the model.

\section{Related work}

\subsubsection{Model-agnostic feature importance.}

Explaining a predictive model on a \emph{global} level aims to understand how important is a given feature to its performance. Several model-agnostic, i.e. explaining any black-box function, feature importance measures have been proposed. The widely adopted approach is permutation feature importance \citep{fisher2019all}, which is also extended to local and partial importance \citep{casaliccio2019visualizing}. \cite{molnar2021relating} introduce confidence intervals for permutation feature importance and \cite{au2022grouped} propose to group features to explain their combined importance. In practice, estimating the marginal importance of a single feature without taking into account correlation structure in data is a challenge \citep{molnar2023model}. To address feature dependence, \cite{watson2021testing} proposes to measure the conditional predictive impact between features and predictions using the knockoff sampling framework; also for categorical features \citep{blesch2023conditional}. Related is work on conditional estimation of Shapley-based feature attributions \cite{aas2021explaining}. As opposed to considering Shapley-based feature importance measures \citep{casaliccio2019visualizing,covert2020understanding}, in this paper, we specifically relate permutation importance to counterfactual explanations.

\subsubsection{Counterfactual explanations.} A counterfactual explanation of model prediction describes the smallest change to the feature values that change the predicted class. It is a useful \emph{local} ``what-if'' explanation for making actionable decisions \cite{wachter2017counterfactual}, \cite{SarantiEtAl:2022:ActionableXAI}. One can optimize to find counterfactuals using model-specific optimization methods, e.g. gradients. Our work relates more to model-agnostic approaches available to any black-box model \citep{karimi2020model}. \cite{dandl2020multiobjective} propose multi-objective counterfactual explanations that take into account data manifold, i.e. how likely it is that the counterfactual data point originates from the training data distribution. \cite{mothilal2020explaining} focus on finding a diverse set of counterfactual data points for a given prediction. Most recent work considers the robustness of such explanations to (potentially adversarial) data perturbations \citep{pawelczyk2023probabilistically}. Contrary to related work, we use the intuition behind counterfactual explanations to construct a novel global explanation of feature importance.


\subsubsection{Domain knowledge in the context of model explanation.} Relating explanations to domain knowledge is an emerging research topic \citep{ema2021}. Crucially, interpreting models should be done with respect to data distribution and its correlation structure \citep{molnar2021relating,baniecki2023grammar}. One idea is to build a surrogate model based on explanations of a black-box function achieving an interpretable predictive interface consistent with domain knowledge \citep{alaa2021machine,gosiewska2021simpler}. Other approaches consider including domain knowledge directly in the algorithm used to learn a predictive function \citep{panigutti2020doctor,confalonieri2021using, pfeifer2022multi}, which is challenging to do in an algorithm-agnostic way. 

In this paper, we introduce feature importance explanation and visualization through counterfactual paths that provide additional information about the graph structure of features. Domain knowledge can both be derived from explanations, as well as participate in estimating more accurate explanations.

\section{Counterfactual paths for explaining black-box models}

\subsubsection{Intuition.}
The herein proposed CPATH algorithm randomly selects paths through the feature space, where each feature represents a node in a fully-connected graph or a user-specified knowledge graph, masking the sampling scheme. Once a path is created, CPATH permutes one feature after the other of the sampled path and terminates as soon as a certain number of class labels swap to the inverse class. We call these paths \textit{counterfactual paths}. From these counterfactual paths, we derive an adjacency matrix weighted by the inferred path lengths. Based on the weighted adjacency matrix the \textit{global} importance of a certain feature is determined. 

\subsection{Mathematical formulation}
Let $\mathcal{M} \colon \mathbb{R}^p \to \{1, \ldots, g\}$, where $g$ is the number of classes, denote the model of interest, which is a classifier. Given an observation $\mathbf{x} = (x_1, x_2, \ldots, x_p) \in \mathbb{R}^p$, the model's prediction is denoted as $\mathcal{M}(\mathbf{x})$.

We describe the problem using weighted directed graphs, (weighted digraphs). A graph $G_{\mathcal{M}} = (V, E, w_{\mathcal{M}})$ is defined as an ordered triple, where $V = {1, \ldots, p}$ represents the set of predictors (explanatory variables) that form the vertices of the graph. The set of directed edges (arcs), denoted as $E$, contains ordered pairs of vertices $(i, j)$ such that there exists a directed edge from vertex $i$ to vertex $j$. The function $w_{\mathcal{M}} \colon E \to \mathbb{R}$ maps the edges to their weights and can also be represented as an adjacency matrix $W$.

The variables used in the $\mathcal{M}$ model which constitute the set of vertices $V$ are predetermined, i.e., these are the variables used in the model fitting process. Similarly, the set $E$ is known -- its form can be derived from a domain knowledge graph describing the process of generating the data. However, if the domain knowledge graph is unknown, it can be assumed that $G_{\mathcal{M}}$ is a complete digraph, meaning that each pair of vertices is connected by a symmetric pair of directed arcs.

The objective is to estimate the function $w_{\mathcal{M}}$ (related to the adjacency matrix $W$) by aggregating \emph{counterfactual paths}  obtained from sampling trajectories of finite-length random walks on the unweighted version of the graph $\widetilde{G_{\mathcal{M}}}$.

\begin{definition}[Counterfactual path] \\
Given a model's predictions $\mathcal{M}(\mathbf{X}) = [\mathcal{M}(\mathbf{x}_1), \ldots, \mathcal{M}(\mathbf{x}_n)] \in \{1, \ldots, g\}^n$ for a~dataset $\mathbf{X} = (\mathbf{x}_1, \ldots, \mathbf{x}_n)^T$, we call a sequence $\mathbf{v} = (v_1, \ldots, v_k)$, $k \leq p$, a counterfactual path under counterfactual policy $\Psi$ if perturbing (e.g., by permuting) values in the features $v_1, \ldots, v_k$ leads to a substantial change in the model's predictions according to the indicator given by $\Psi$, i.e, if 
$$\Psi(\mathcal{M}(\mathbf{X}), \mathcal{M}(\mathbf{X}'_{\mathbf{v}})) = 1.$$
\end{definition}

In the above definition, $\mathbf{X}'_{\mathbf{v}}$ represents a dataset with perturbed values of features from the sequence $\mathbf{v}$ (note that perturbations are performed sequentially, for individual variables as they are added to the path). The counterfactual policy $\Psi$ is used to determine whether a perturbation of features leads to a significant change in predictions.

Various choices for the counterfactual policy $\Psi$ are possible. One approach is to define $\Psi$ based on the fraction of changed predictions $p$, where 
$$  p = \frac{1}{n} \sum_{i=1}^n \mathbbm{1}(\mathcal{M}(\mathbf{x}_i) \neq \mathcal{M}(\mathbf{x}'_i)).
$$  
With that, we define $\Psi$  in a stochastic way as follows:
$$
    \Psi_{\mathrm{s}}(\mathcal{M}(\mathbf{X}), \mathcal{M}(\mathbf{X}'_{\mathbf{v}})) \coloneqq X \sim Bern(p).
$$
Alternatively, $\Psi$ can be defined as an indicator of whether the fraction of changed predictions exceeds the selected threshold $\kappa$ chosen by a user, i.e., 
$$\Psi_{\mathrm{d}}(\mathcal{M}(\mathbf{X}), \mathcal{M}(\mathbf{X}'_{\mathbf{v}})) \coloneqq \mathbbm{1}(p > \kappa).$$
Another option is to penalize path lengths by considering their length $k$ in the definition of $\Psi$.

The selection of an appropriate counterfactual policy $\Psi$ is a crucial step in the proposed methodology. Once the counterfactual policy is defined, Algorithm \ref{alg:cpaths-generation} is employed to identify and collect counterfactual paths. This algorithm takes several inputs, including the model of interest $\mathcal{M}$, the dataset $\mathbf{X}$, the counterfactual policy $\Psi$, the unweighted graph $\widetilde{G_{\mathcal{M}}}$, the number of iterations $n_{iter}$, and the maximal path length $k$. The CPATH algorithm initializes a set to store counterfactual paths and iterates a specified number of times. In each iteration, a new path is generated by sampling vertices from the unweighted graph and checking for significant changes in the model's predictions based on the counterfactual policy.

\SetKwComment{Comment}{/* }{ */}
\begin{center}
\begin{algorithm}[H]
\SetAlgoLined
\KwData{$\mathcal{M}$, $\mathbf{X}$, $\Psi$, $\widetilde{G_{\mathcal{M}}}$, $n\_iter \in \mathbf{Z_+}$, $k \in \mathbf{Z_+}$}
$CPATHS \gets \{\}$\;
\For{$i=1$ \KwTo $n\_iter$}{
    $\mathbf{v} \gets [\,]$\;
    \For{$j=1$ \KwTo $k$}{
    $v_j \gets \mathrm{sample\_vertex}(\widetilde{G_{\mathcal{M}}})$\;
    $\mathbf{v} \gets \mathbf{v} + [v_j]$\;
    \If{$\Psi(\mathcal{M}(\mathbf{X}), \mathcal{M}(\mathbf{X}'_{\mathbf{v}})) = 1$}{
        $CPATHS \gets CPATHS \cup \{\mathbf{v}\}$ \;
        \textbf{break}\;
    }
    }
}
\caption{CPATH: counterfactual paths generation}
\label{alg:cpaths-generation}
\end{algorithm}
\end{center}

Subsequently, the observed counterfactual paths are used in Algorithm \ref{alg:transition-matrix-generation} to estimate the function $w_{\mathcal{M}}$, i.e., derive the corresponding adjacency matrix $W$. Specifically, the transition matrix $\mathbf{T}$ is calculated by iterating through the generated paths. For each path, the length $l$ of the path is computed. Then, for each consecutive pair of vertices $(v_i, v_{i+1})$ in the path, the corresponding entry in $\mathbf{T}$ is incremented by $k - l + 1$. This process accounts for the penalization of longer paths, as paths with a shorter length contribute more to the edge weights. The resulting adjacency matrix should capture the underlying dependencies and interactions between variables based on the observed counterfactual paths.

\begin{center}
\begin{algorithm}[H]
\SetAlgoLined
\KwData{$CPATHS$, $k \in \mathbf{Z_+}$}
$\mathbf{T} \gets \mathbf{0}_{p \times p}$\;
\For{$\mathbf{v}$ $\mathrm{\mathbf{in}}$ $CPATHS$}{
    $l \gets \mathrm{length}(\mathbf{v})$\;
    \If{$l = 1$}{
         $\mathbf{T}[v_1, v_1] \gets \mathbf{T}[v_1, v_1] + k - l + 1 $\; 
    }
    \For{ $i=1$ \KwTo $l - 1$}{
        $\mathbf{T}[v_i, v_{i+1}] \gets \mathbf{T}[v_i, v_{i+1}] + k - l + 1 $\; 
    }
}
\caption{CPATH: transition matrix generation}
\label{alg:transition-matrix-generation}
\end{algorithm}
\end{center}

Based on the determined adjacency matrix $W$, we can calculate the importance of individual variables. A straightforward approach is to compute the fraction of weights of edges adjacent to the node corresponding to variable $\mathbf{X}_j$. This can be expressed as:
\begin{equation}
    Imp(\mathbf{X}_j) = \frac{\sum_{i\in [p]} \mathbf{T}[v_i,v_j]}{\sum_{i\in [p], k\in [p]} \mathbf{T}[v_i,v_k]},
\end{equation}
where $\mathbf{T}$ is the transition matrix obtained from Algorithm \ref{alg:transition-matrix-generation} and $[p]$ denotes the set of indices from $1$ to $p$.

Another way to estimate the importance values is by considering the stationary distribution of a Markov chain based on transition matrix $\mathbf{P}$ being the row-wise normalized matrix $\mathbf{T}$. The stationary distribution represents the long-term probability distribution of being at each node in the graph and it is a vector $\pi$ that satisfies the condition $\pi \mathbf{P} = \pi$. In this setting, the importance of variable $\mathbf{X}_j$ is given by the corresponding element $\pi_j$ in the stationary distribution vector.

\subsection{Counterfactual paths for causal modeling}
In the previous section the transition matrix $\mathbf{T}$ was used to derive feature importances of single features. However, it also contains information about important interactions of features in form of a weighted directed graph.  
In fact, the generated counterfactual paths in $CPATHS$ can be used for causal modeling and inference, for instance with Bayesian networks. From the absence and presence of features within the detected counterfactual paths one can estimate the conditional probability distribution between the important features, which may help to discovery causal effects. We illustrate this on two concrete examples in Section~\ref{sec:results}.

\section{Evaluation and experimental set-up}

\subsubsection{General approach.}
We have generated synthetic data under three conditions: Conditional dependency, correlation, and conditional independence (see Appendix section A.1). For each of these datasets two features were simulated fulfilling the above-mentioned conditions; the rest of the features were added as noise. A Random Forest was trained on the generated synthetic data and predictions were made based on the training set. The model was treated as a black box and explainable methods were exploited to verify the relevant features. 

\subsection{Explaining the model}\label{sec:explaining-the-model}
Our evaluation strategy was to compare the feature importance scores computed by the explainers (e.g. SHAP~\citep{aas2021explaining} and LIME~\citep{lime}) with the model-intern Gini impurity scores generated by the random forest, which we consider here as ground truth. A high correlation between the explainers' feature importance scores and the model's Gini impurity values indicates that the explainer can detect the underlying patterns in the model's behaviour. Here, we report on the correlation with the model-specific Gini impurity scores associated with each simulated feature. We compare our approach with SHAP, LIME, and Permutation Feature Importance~\citep[PFI,][]{fisher2019all}. We report and analyze the performance of PFI, because it relies on a permutation scheme to derive feature importances, and thus is related to our method. Unlike CPATH, however, PFI requires the ground-truth labels to determine feature importance. 

\subsection{Explaining the data}
We are aware that the Gini impurity scores themselves are biased due to several shortcomings \citep{nembrini2018revival}. As a consequence, 
we also studied the capability of the explainers to detect the most important features within the data. We could show that the model-specific Gini impurity scores efficiently reflect the simulated ground truth, which essentially supports the validity of the first evaluation set-up as described in Section~\ref{sec:explaining-the-model}. In this particular investigation, however, the focus was more on the simulated features within the data and not on the features the model actually preferred, which is addressed by our first evaluation (see Section~\ref{sec:explaining-the-model}). 

In the case of SHAP and LIME, feature importance scores for \textit{global} explanations were computed as the mean absolute values of the local explanation scores. For this data-centric experiment we compared our method also to Conditional Predictive Impact \citep[CPI,][]{watson2021testing}. CPI provides variable importance measures taking into account the association between one or several features and a given outcome.

\subsection{Evaluation of the explanations' quality}

There are several methods that are used to measure the quality of explanations; in a survey of surveys \citep{Schwalbe:2023:ComprehensiveTaxonomyxAI} describe a taxonomy of Explainable AI methods and mention the fundamental differences and commonalities between several quality of explanation metrics. Expressing the expectation of the fact that a substantial change in the model decision's logic will also be reflected in its explanations is made by the sensitivity metric. Furthermore, fidelity (also referred to as faithfulness) is mainly used in conjecture with a surrogate model \citep{Ancona:2017:UnderstandingGBAttribution}. It describes the agreement between the original and surrogate model that is used to provide the explanations. Typically, models strive to have low infidelity, which is computed by the following function:

\begin{equation}
\textrm{INFD} ( \mathrm{\Phi}, \mathbf{f}, \mathbf{x}) = \mathbb{E}_{\mathbf{I} \sim \mu_{\mathbf{I}}} \biggl[ \mathbf{I}^T \mathrm{\Phi}(\mathbf{f}, \mathbf{x}) - \Bigl( \mathbf{f}(\mathbf{x}) - \mathbf{f}(\mathbf{x} - \mathbf{I}) \Bigl)^2 \biggr]
\label{eq:equation_infidelity}
\end{equation}

Equation \ref{eq:equation_infidelity} provides the infidelity for a model computing the function $\mathbf{f}$, having $\mathbf{x}$ as input. The explanation method is represented by $\mathrm{\Phi}$, called ``explanation functional''. This metric is based on the definition of suitable perturbations, tailored to the needs of the task, the used model, input and the xAI method. They are represented by the difference $\mathbf{I} = \mathbf{x} - \mathbf{x}_0 \in \mathbb{R}^d$ between the input $\mathbf{x}$ and $\mathbf{x}_0$, where $\mathbf{x}_0$ can be a baseline value, a noisy baseline value or even a Random Variable (RV) (in the equation \ref{eq:equation_infidelity}, $\mathbf{I} \sim \mu_{\mathbf{I}}$ is a RV having the probability measure~$\mu_{\mathbf{I}}$). 

Another metric that is related to infidelity to a certain extent, since it is also based on the principle of using perturbations is sensitivity. In this case, it is not the input data $\mathbf{x}$ that is perturbed but the input features; the perturbation strategy is not defined solely by the data scientist, but it is driven by the relevance/attribution values that each feature has - according to the xAI method of interest. This metric expresses an expectation that the computed attribution values for each feature have some relationship with the computed output of the model. If a feature that is said to be highly important or decisive for a particular prediction is removed or replaced by a less informative one, then this action has to have some impact on the model's prediction. Even more, this change has to be in some way analogous to the assigned relevance/attribution of the feature, as computed by the xAI method. The extension of sensitivity to subsets of features (and not just one) is called sensitivity-$n$, where $n$ is the cardinality of the selected feature subset $S$. The corresponding equation is:

\begin{equation}
\text{Sens}_n = r \Big( \sum_{s \in S_i} \mathbf{e}_s, \mathbf{f} (\mathbf{x})_c - \mathbf{f} (\mathbf{x}_{S_i})_c \Big)
\end{equation}

where $c$ is the predicted class, $\mathbf{x}$ is the original input, $\mathbf{x}_S$ is $\mathbf{x}$ with all features in subset $S$ removed. The sum of attributions in the subset $S$ is denoted by $\sum_{s \in S_i} \mathbf{e}_s$ and its Pearson correlation $r$ with the aforementioned difference comprises the value of sensitivity-$n$. What is questionable is the use of this correlation without considering Spearman's rank correlation coefficient or Mutual Information (MI) \citep{MacKay:2003:InformationTheoryInferenceBook} for capturing non-linear correlations.

A detailed description of the relationship between sensitivity and infidelity for different scenarios is described in \citep{Yeh:2019:OnTheFidelitySensitivityOfExplanations}; the researchers have shown that a small decrease in sensitivity can be achieved when the explanation is multiplied by a smooth kernel (in this case a Gaussian) without increasing the infidelity - and under particular circumstances also decreasing it. This provided a general methodology that can start from any explanation towards one that has more beneficial quality properties - at least as far as those metrics are concerned. A detailed analysis of several related metrics with different representative experiments testing various datasets in the image processing domain is presented in \citep{Gevaert:2022:EvaluatingFeatureAttributionImageDomain}.  


\subsection{Explanations based on domain knowledge graphs}

In the above-described experiments, the counterfactual paths are computed on a fully-connected graph. The domain expert, however, might have some prior knowledge about the functional relationships between the studied features (nodes in the graph), which could be reflected as edges in a knowledge graph. In these cases, the explanatory factors induced by the counterfactual paths and their visiting nodes are restricted and guided by domain knowledge.
\\
In this specific evaluation, we compute the network, features and classes in the following way. First, we generate Barabasi networks \citep{barabasi1999emergence} of varying size and structure. Second, we compute the features associated with a node using a normal distribution with $N(\mu = 0, \sigma = 1)$. In a next step, we randomly select two connected nodes $ v_{1}$ and  $v_{2}$ and apply the following formula

\begin{equation*}
    z = 5 * v_{1} + 3 * v_{2}.
\end{equation*}

We apply the sigmoid function to transform $z$ into a range between 0 and 1

\begin{equation*}
    S(z) = \frac{1}{1 + e^{-z}}.
\end{equation*}

After we get the transformed values from the sigmoid function, we underlay a Bernoulli distribution and sample from it to obtain the outcome class. 

In the described experiment the feature sampling in CPATH is guided by the network through random walks. In the remainder of this paper, we refer to it as CPATH$_{know}$.

Finally, we exemplary show the potential of our approaches (CPATH and  CPATH$_{know}$) in a bioinformatics application for biomarker discovery. We used the gene expression data of human breast cancer patient samples for an experimental evaluation of the herein proposed methodologies. The data was retrieved from The Cancer Genome Atlas (TCGA) and was preprocessed as described in \citep{chereda2021explaining}. We masked the data by the topology of the Human Protein Reference Database (HPRD) protein-protein interaction (PPI) network \citep{keshava2009human}. The resulting dataset comprised 981 patients and 8469 genes. The binary prediction task was to classify the samples into a group of patients with the luminal A subtype (499 samples) and patients with other breast cancer subtypes (482 samples). Knowledge-guided explanations were generated using CPATH$_{know}$ for the detection of potential breast cancer-specific biomarkers.

\section{Results}\label{sec:results}

\subsection{Synthetic data: correlation with ground-truth}
\begin{figure}
     \centering
         \includegraphics[width=\textwidth]{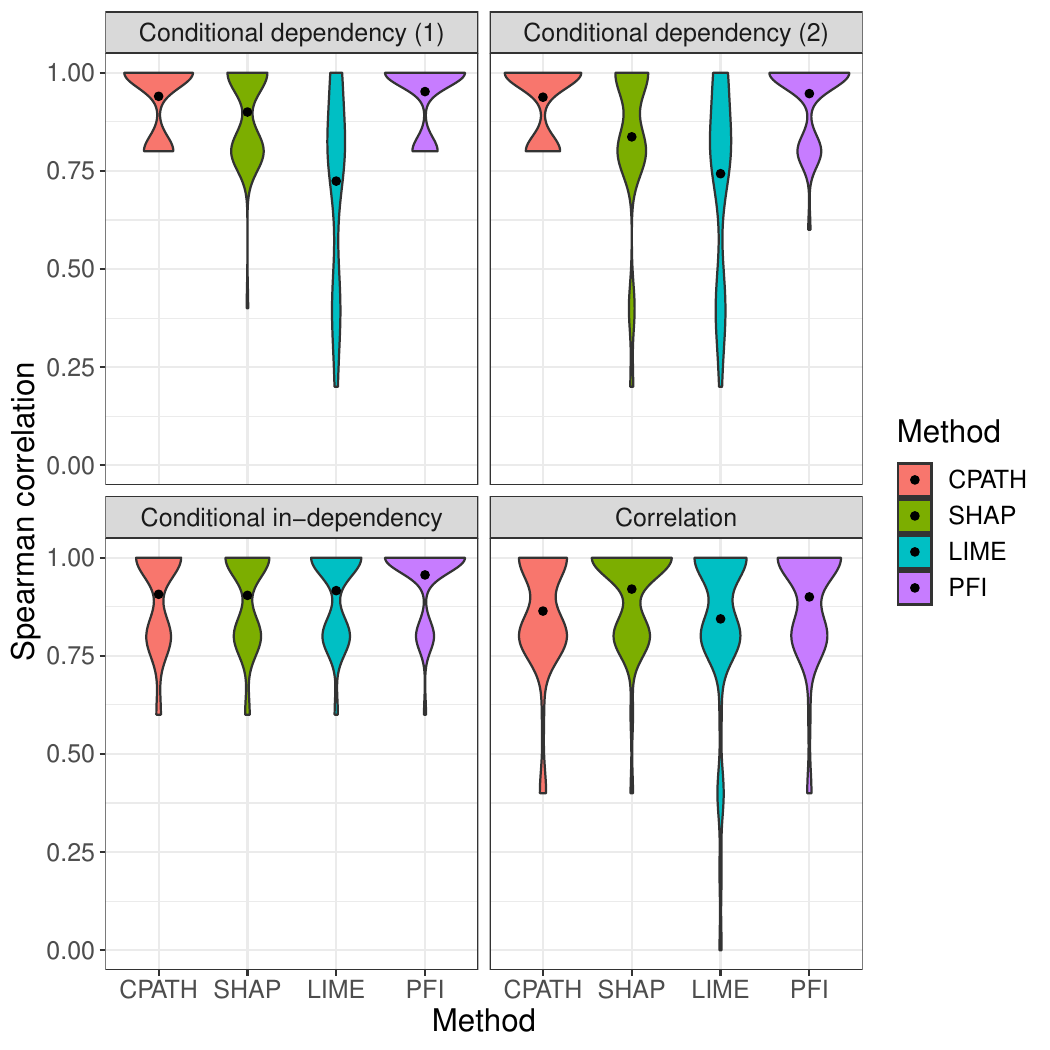}
         \caption{Correlation with Gini importance values based on 50 simulations.}
        \label{fig:sim_corr}
\end{figure}

\begin{figure}
     \centering
         \includegraphics[width=\textwidth]{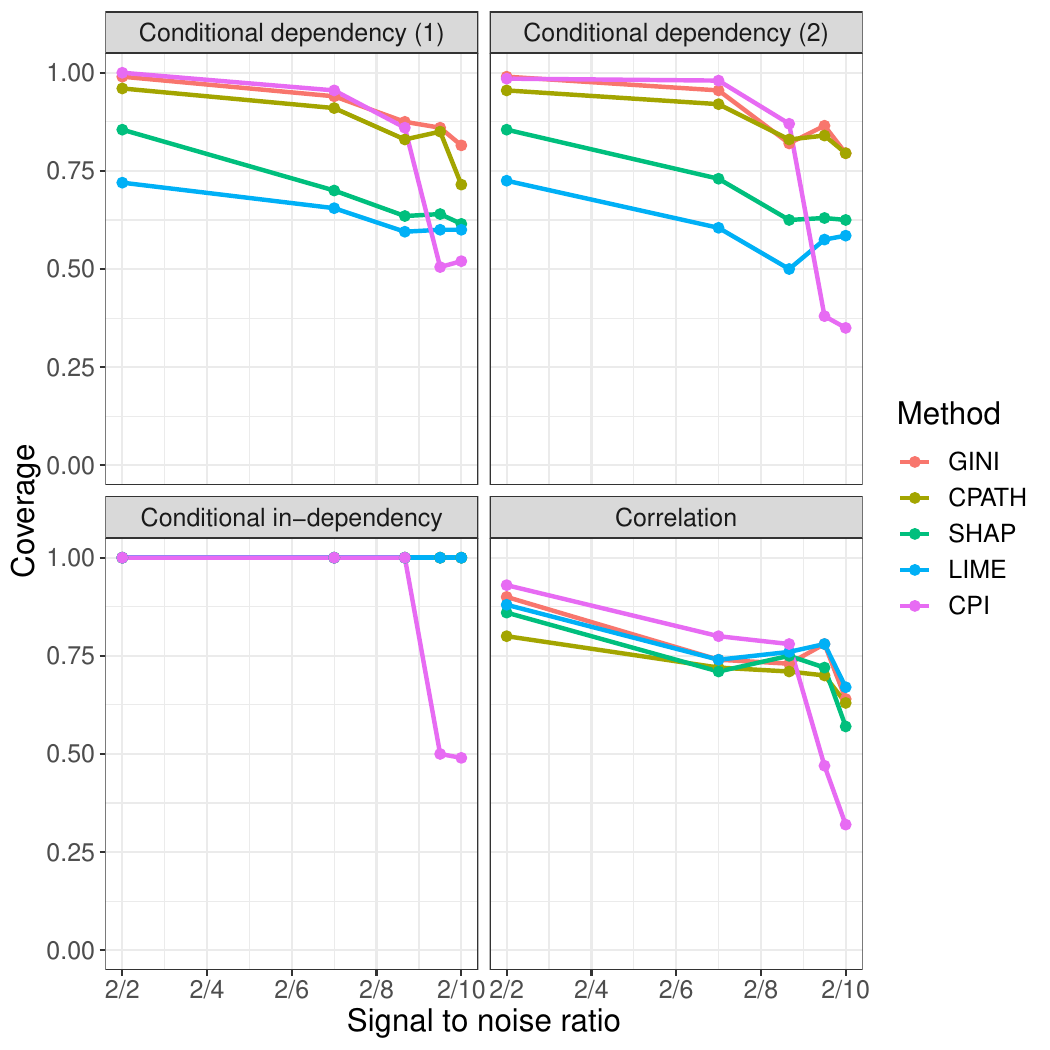}
         \caption{Mean coverage of important features within the simulated data based on 50 simulations and along the signal-to-noise ratio.}
        \label{fig:sim_cov}
\end{figure}

The correlation with the Gini impurities (herein assumed as the \textit{model-specific} \textit{ground-truth}) is depicted in Figure~\ref{fig:sim_corr}. The results indicate that path-based explanations (CPATH) and permutation-based feature importances (PFI) are more accurate compared to SHAP and LIME. Especially in the case of conditional feature dependency they efficiently reflect the model's internal behaviour (see Figure~\ref{fig:sim_corr}~\&~Table~\ref{tab:cor_gini}). 
The explanations based on LIME are not informative. The same observation can be made when interpreting the ability of the methods to detect the relevant features within the data (see Figure~\ref{fig:sim_cov} and Table \ref{tab:cov}). Here, the simulated features are defined as the \textit{data-specific} \textit{ground-truth}. In the case of conditional feature dependency, CPATH is clearly more accurate than SHAP and LIME. CPATH almost perfectly aligns with the ground truth. CPI outperforms all methods when the signal-to-noise ratio is low. However, there is a substantial performance degradation when this ratio increases. It should be noted that CPI is specially designed for feature selection purposes and not for explaining the internals of the model. For instance, specialized feature selection methods often use a random forest, or any other classifier as a wrapper to detect the most relevant features within the data \citep{Pfeifer:2022:FeatureTrustAI}. 

In the case of correlated features, CPATH performs slightly worse than the competing SHAP method (see Figure~\ref{fig:sim_corr}). Also Figure \ref{fig:sim_cov} suggests that SHAP is the more appropriate explainer in the presence of correlated features. In this data-focused experiment (Figure \ref{fig:sim_cov}, Table \ref{tab:cov}) also LIME is closer to the \textit{ground-truth}. Notably, when the signal-to-noise ratio is high all methods perform almost identically.

\begin{table}[ht]
\centering
\caption{Correlation to Gini importances (min/mean/max)}
\vspace{1em}
\begin{tabular}{l c c  c  c  c}
  \toprule
  \textbf{Experiment} & & \textbf{CPATH} & \textbf{SHAP} & \textbf{LIME} & \textbf{PFI} \\
  \midrule			
  Conditional dependency (1)   && 0.8/0.94/1 &  0.4/0.90/1 & 0.2/0.72/1 & 0.8/0.95/1 \\
  Conditional dependency (2)   && 0.8/0.94/1 & 0.2/0.84/1  & 0.2/0.74/1 & 0.6/0.94/1  \\
  Correlation && 0.4/0.87/1 &  0.4/0.92/1 &  0/0.84/1  & 0.4/0.90/1 \\
  Conditional in-dependency && 0.6/0.91/1 &  0.6/0.90/1 & 0.6/0.92/1 & 0.6/0.95/1\\
  \bottomrule  
\end{tabular}
\label{tab:cor_gini}
\end{table}

\begin{table}[ht]
\centering
\caption{Mean coverage of the important features within the data}
\vspace{1em}
\begin{tabular}{l c c c c c c }
  \toprule
  \textbf{Experiment} & \textbf{signal/noise} & \textbf{GINI} & \textbf{CPATH} & \textbf{SHAP} & \textbf{LIME} & \textbf{CPI} \\
  \midrule			
  Conditional dependency (1) & 2/2 & 0.990 & 0.960 &  0.855 & 0.720 & 1  \\
  Conditional dependency (2) & 2/2 & 0.990 & 0.950 & 0.905  & 0.730  & 0.995  \\
  Correlation & 2/2 & 0.995 & 0.960 &  0.905 & 0.725 & 1  \\
  Conditional in-dependency & 2/2 & 1 & 1 &  1 & 1 & 1  \\
  \midrule
  Conditional dependency (1) & 2/4 & 0.940  & 0.910  & 0.700  & 0.655 & 0.955  \\
  Conditional dependency (2) & 2/4 & 0.940 & 0.880 & 0.710  & 0.665  & 0.925  \\
  Correlation & 2/4  & 0.760 & 0.700  & 0.705  & 0.770  & 0.805   \\
  Conditional in-dependency & 2/4 & 1 & 1 & 1  & 1 & 1  \\
  \midrule
  Conditional dependency (1)& 2/6 & 0.875 & 0.830  & 0.635 & 0.595 & 0.860  \\
  Conditional dependency (2) & 2/6 & 0.870 & 0.820 & 0.670  & 0.655 & 0.895  \\
  Correlation & 2/6  & 0.730 & 0.745 & 0.675 & 0.770 & 0.775  \\
  Conditional in-dependency & 2/6 & 1 & 1 & 1  & 1 & 1  \\
  \midrule
  Conditional dependency (1)& 2/8 & 0.860 & 0.850  & 0.640  & 0.600 & 0.505  \\
  Conditional dependency (2) & 2/8 & 0.845  & 0.805 & 0.615  & 0.600 & 0.400  \\
  Correlation & 2/8 & 0.685 & 0.710 & 0.660 & 0.755 & 0.415  \\
  Conditional in-dependency & 2/8 & 1  & 1 & 1  & 1 & 0.500  \\
  \midrule  
  Conditional dependency (1) & 2/10 & 0.815 & 0.715  & 0.615  & 0.600 & 0.520  \\
  Conditional dependency (2) & 2/10 & 0.820 & 0.735 & 0.585  & 0.575 & 0.325   \\
  Correlation & 2/10  & 0.645 & 0.670 & 0.640 & 0.655 & 0.390  \\
  Conditional in-dependency & 2/10 & 0.995 & 0.900 & 0.995  & 0.995  & 0.495  \\
  \bottomrule
\end{tabular}
\label{tab:cov}
\end{table}

\subsection{Sensitivity, Infidelity, and Interpretability}

We have evaluated the sensitivity and infidelity of the explanations based on four different datasets, namely Ionosphere, Breast Cancer, Diabetes, and Iris. The datasets were retrieved using the R package mlbench \citep{mlbench}. We split the data into a train (80\%) and test set (20\%) and generated explanations based on the test set. In terms of sensitivity, CPATH performs well on all datasets and outperforms LIME and SHAP on two out of four cases (see~\ref{fig:sensitivity}). LIME in this experiment is the worst-performing method. The fidelity of the explanations based on CPATH is not as good as SHAP, but it is competitive with LIME (see~\ref{fig:fidel}). 

\begin{figure}[ht]
  \centering
  \includegraphics[scale=0.67]{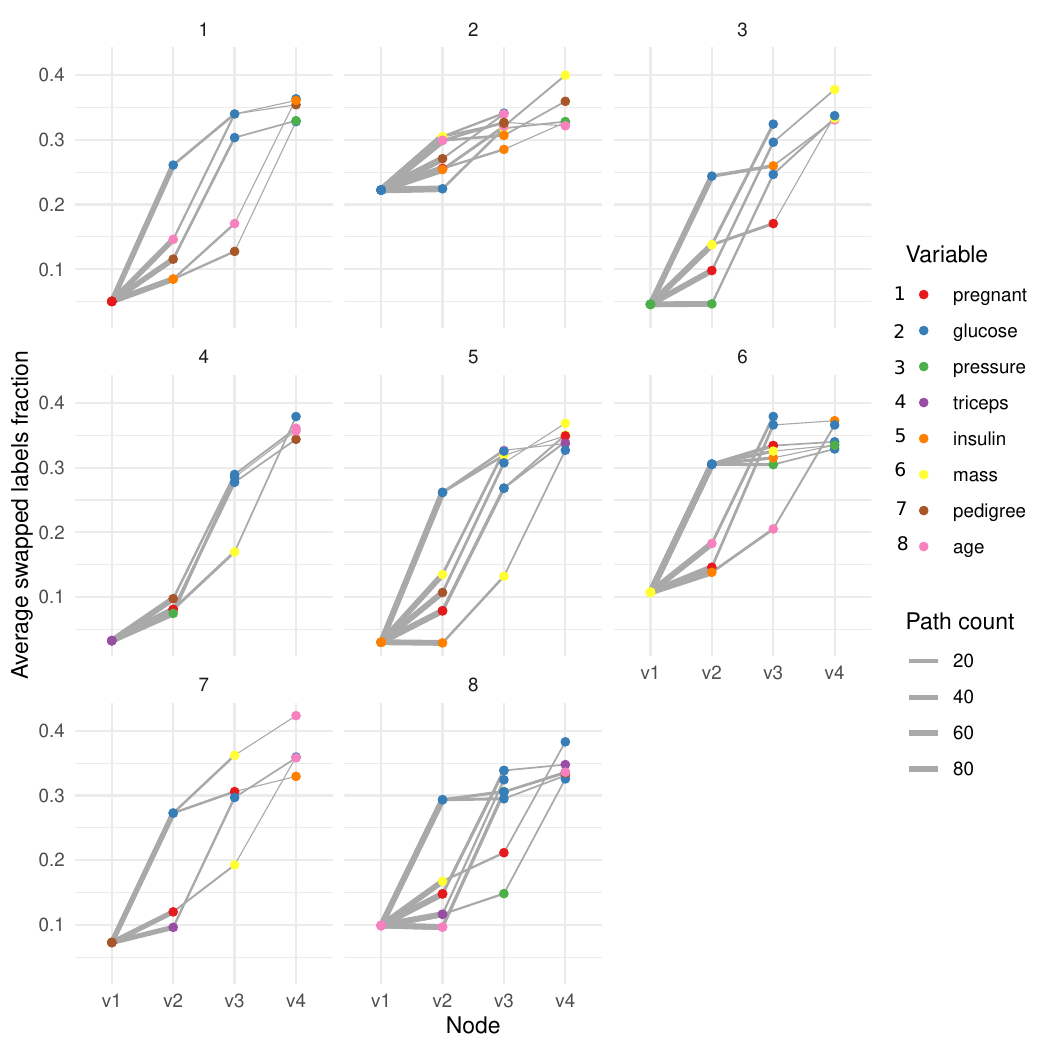}
  \caption{Diabetes dataset. Graphical summary of the counterfactual paths.}
   \label{fig:DIAB_appl}
\end{figure}

\begin{figure}
     \centering
     \begin{subfigure}[b]{0.49\textwidth}
         \centering
         \includegraphics[width=\textwidth]{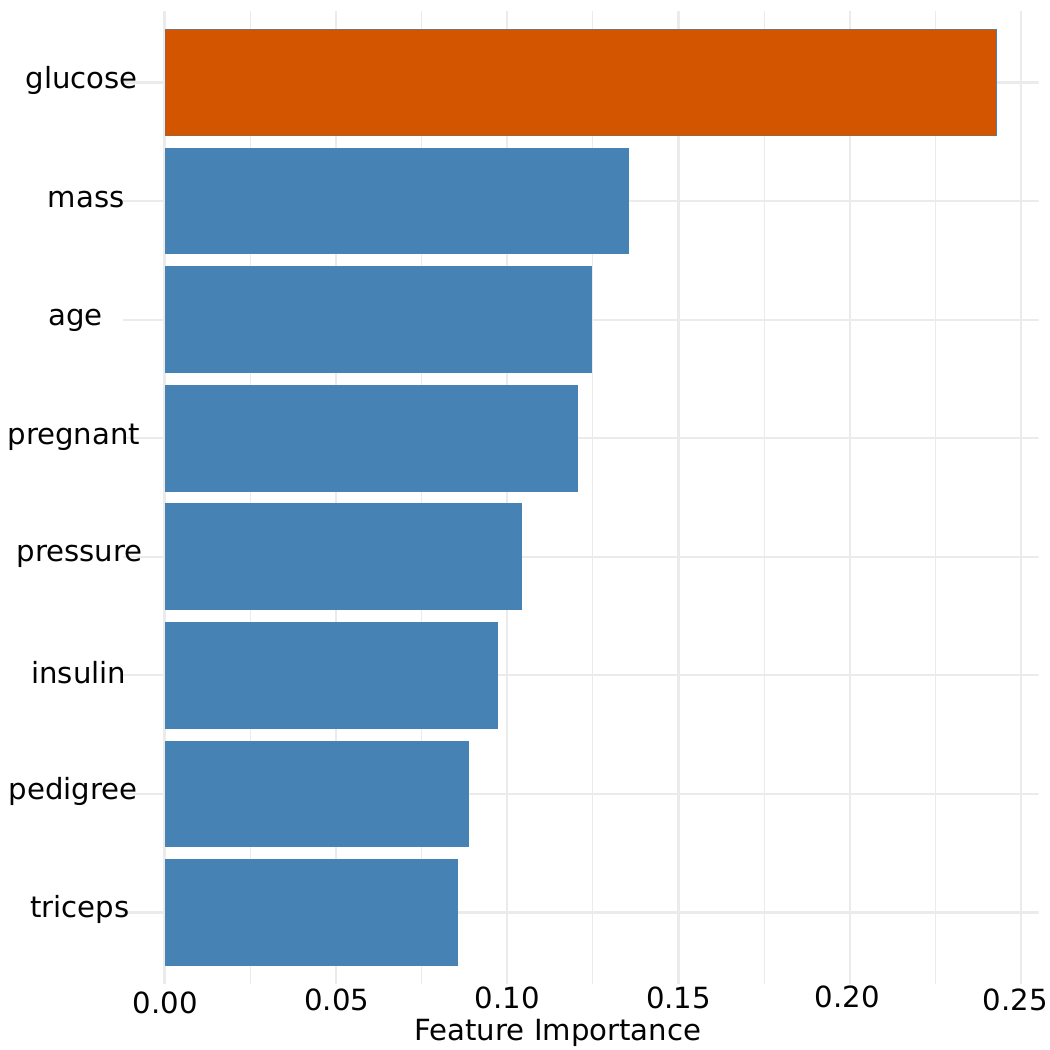}
         \caption{Feature importance}
     \end{subfigure}
     \begin{subfigure}[b]{0.50\textwidth}
         \centering
         \includegraphics[width=\textwidth]{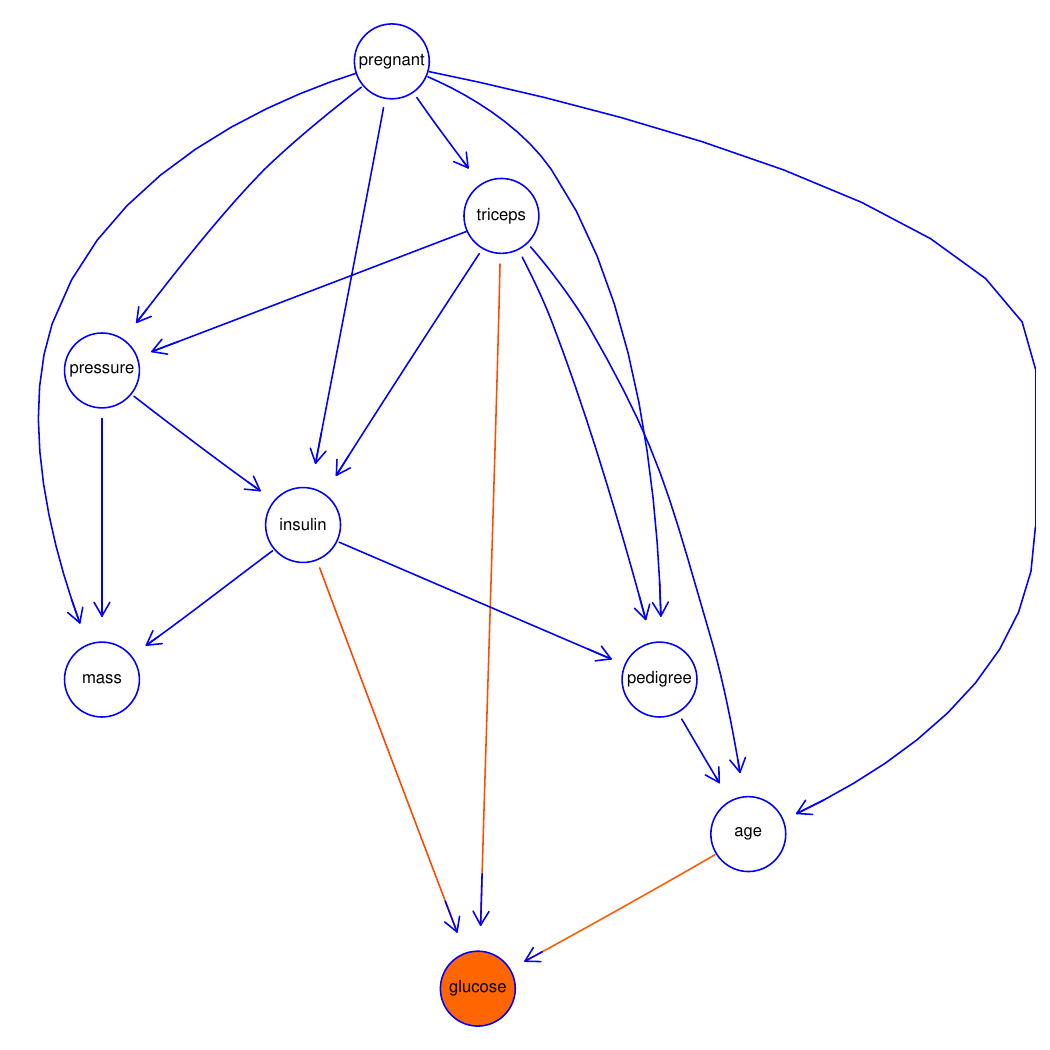}
         \caption{Conditional dependencies}
     \end{subfigure}
        \caption{
      Diabetes dataset. Feature importances and conditional dependencies inferred by Bayesian network learning and here represented as a Directed Acyclic Graph.
       }
        \label{fig:DIAB_appl_fi}
\end{figure}

To showcase the interpretability of our proposed approach we analyzed the Diabetes dataset in more detail (see Figure~\ref{fig:DIAB_appl}). The accuracy of the random forest classifier was $AUC=0.97$. A graphical summary of the generated counterfactual is shown in Figure~\ref{fig:DIAB_appl}. We see that the glucose variable causes the largest number of swapped labels when it is used as the first node in a counterfactual path. More than 20\% labels on average swap when this feature is permuted. Once permuted, insulin and mass increase the swapped fraction up to 30\%. However, the largest fraction is obtained by the path starting with pedigree and further going through glucose, mass, and age. This particular path caused $>$ 40\% of swapped labels on average. These observations suggest that a combination of the mentioned variables might be a marker for a detailed medical risk assessment. The feature importances derived from the counterfactual paths can be obtained from Figure \ref{fig:DIAB_appl_fi}. We further learned a Bayesian network from the inferred counterfactual paths (Figure \ref{fig:DIAB_appl_fi}), using the R-package \texttt{bnlearn} \citep{bnlearn}. While the glucose variable was inferred as the most important feature, the Directed Acyclic Graph in Figure~\ref{fig:DIAB_appl_fi}b indicates that it depends on several features, including age, triceps and insulin, causing the overall importance. From the learned conditional distributions we see that when these features are present within the given path, it increases the probability of glucose up to $0.72$ to be also part of that path, causing the label swap. 

\subsection{Knowledge-guided explanations and application on Protein-Protein Interaction Network (PPI)}

\subsubsection{Syntethic Barabasi networks.}
\begin{figure}
     \centering
     \begin{subfigure}[b]{0.49\textwidth}
         \centering
         \includegraphics[width=\textwidth]{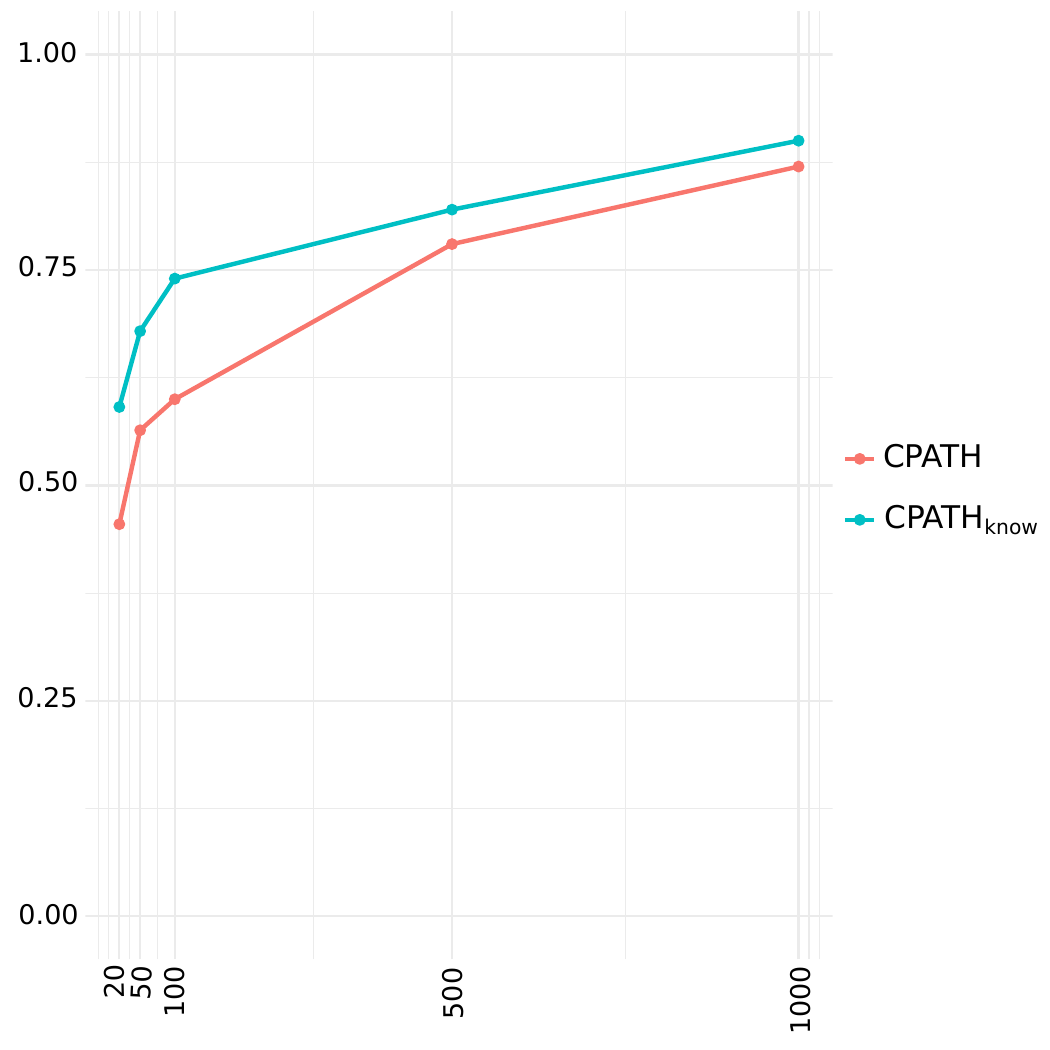}
         \caption{Varying number of paths}
     \end{subfigure}
     \begin{subfigure}[b]{0.49\textwidth}
         \centering
         \includegraphics[width=\textwidth]{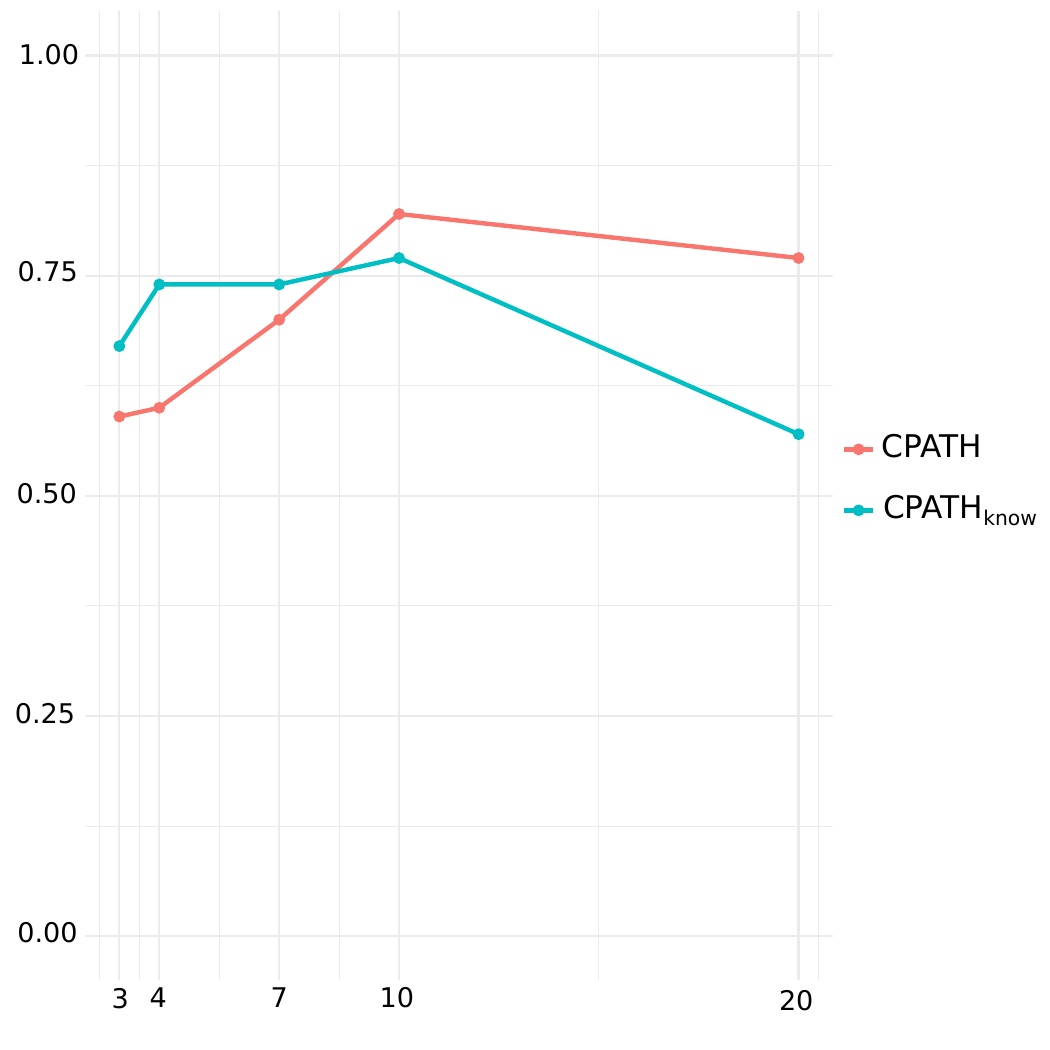}
         \caption{Varying length of paths}
     \end{subfigure}
        \caption{
       Mean coverage in detecting the  important features within the simulated data, masked by Barabasi networks. The results are based on 100 simulations, where data and network topology vary in each iteration. (\textbf{a}) The path length is set to $k=4$. (\textbf{b})~The number of paths is set to $100$.
       }
        \label{fig:know}
\end{figure}

The experiments on synthetic Barabasi networks indicate that incorporated domain knowledge generates more parsimony and interpretable explanations (see Figure~\ref{fig:know}). CPATH with incorporated knowledge is more accurate in detecting the relevant features when the path length is low. CPATH without guided domain knowledge requires about three times larger paths to capture the ground truth. Furthermore, knowledge-guided counterfactual explanations converge faster, as indicated by the higher performance when the number of sampled paths is low (Figure~\ref{fig:know}a). Interestingly, CPATH without domain knowledge significantly outperforms CPATH$_{know}$ when the size of the sampled paths is high. In that case, CPATH is able to explore the feature space more efficiently, while the incorporated knowledge graph seems to hinder optimal convergence. The reason for that behavior is due to the edge degrees of the graph which leads to repeated visits of the same node in a random walk. Thus, we can derive the conclusion that for incorporated knowledge, in the form of a graph, a high number of generated paths is essential to ensure a sufficient number of starting nodes so that the whole graph can be explored by the random walks.

\subsubsection{Application on PPI networks.}
\begin{figure}
     \centering
     \begin{subfigure}[b]{0.50\textwidth}
         \centering
         \includegraphics[width=\textwidth]{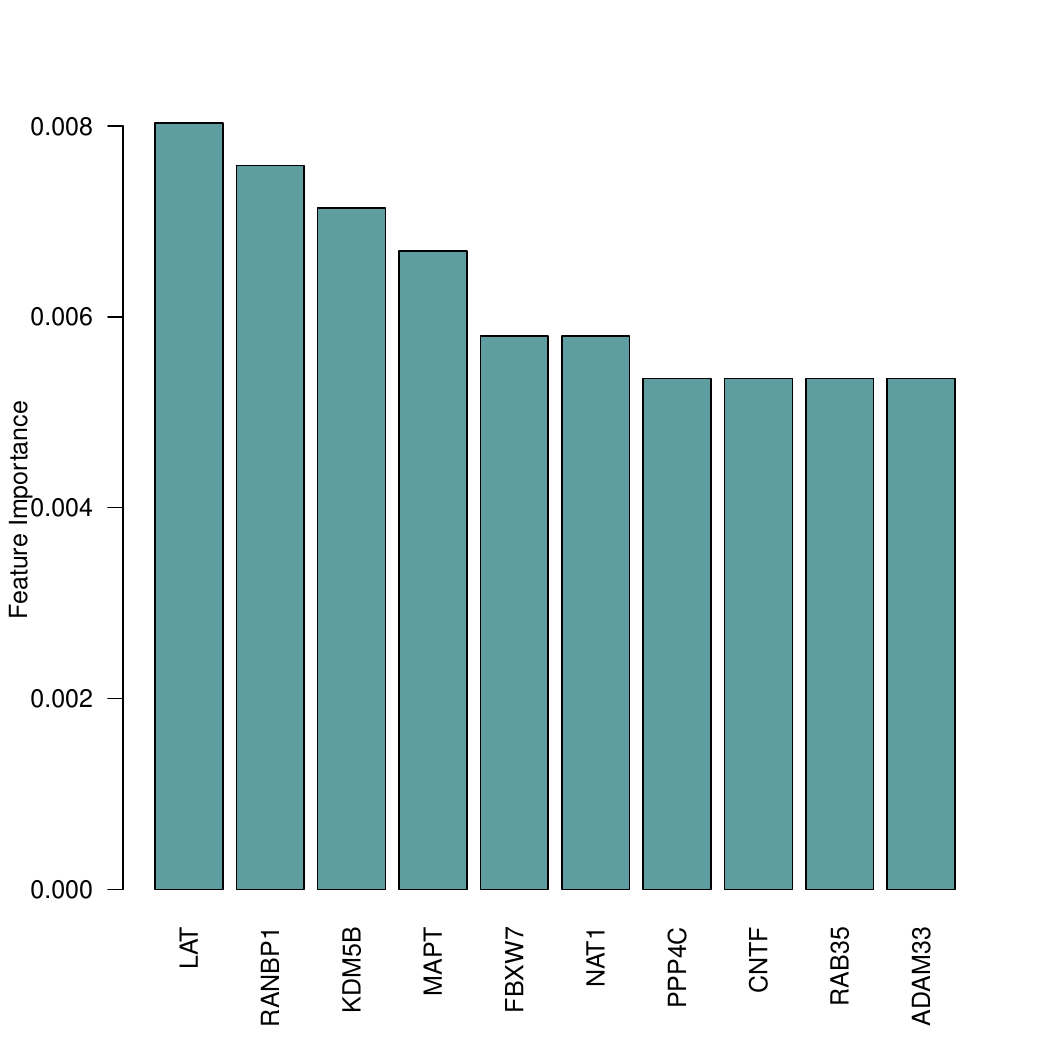}
         \caption{Top-10 relevant genes}
     \end{subfigure}
     \begin{subfigure}[b]{0.49\textwidth}
         \centering
         \includegraphics[width=\textwidth]{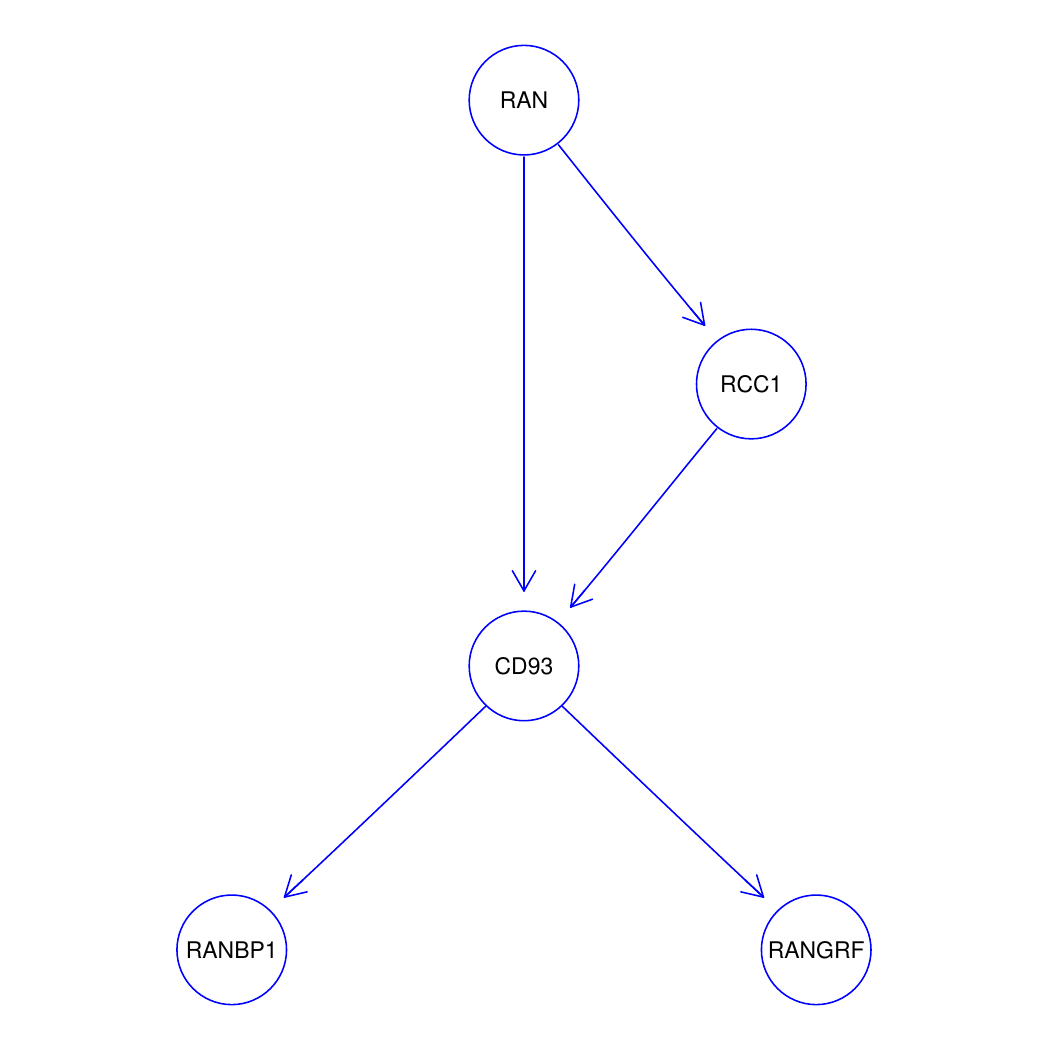}
         \caption{Conditional dependencies}
     \end{subfigure}
        \caption{
      Breast cancer dataset. (\textbf{a}) Top-10 relevant genes inferred by CPATH with incorporated PPI domain knowledge. (\textbf{b}) Conditional dependencies of the neighborhood features of RANBP1 inferred by Bayesian network learning and here represented as a Directed Acyclic Graph.
       }
        \label{fig:PPI1}
\end{figure}

The retrieved breast cancer gene expression data was split into a train (80\%) and test set (20\%). We trained a random forest comprising $1000$ trees and we evaluated the accuracy based on the independent hold-out test set. The trained classifier achieved an Area Under the ROC Curve (AUC) of $0.93$. Knowledge-guided explanations were generated using CPATH${_{know}}$ using a PPI network. We have generated $10000$ paths using a path length of $k=10$. Overall, $118$ paths of the generated paths were counterfactuals. The feature importances derived from these paths can be obtained from Figure~\ref{fig:PPI1}. The top-3 genes were LAT, RANBP1, and KDM5B. Single-feature test set accuracy for these genes were $\text{AUC(LAT)}=0.49$, $\text{AUC(RANBP1)}=0.73$, and $\text{AUC(KDM5B)} = 0.54$. These results suggest that LAT and KDM5B on their own are not sufficient as predictive markers. 

LAT is a protein that plays a critical role in the immune system and is primarily involved in T cell activation and signaling. It is expressed in T cells and other immune cells, facilitating the transmission of signals from the T cell receptor to the downstream signalling molecules.

The KDM5B gene, also known as JARID1B or PLU-1, is a gene that encodes a protein belonging to the family of histone demethylases. Histone demethylases are enzymes involved in the regulation of gene expression by modifying histone proteins, which are involved in packaging DNA within the nucleus. Studies have implicated KDM5B in various biological processes, including development, differentiation, and cancer. While KDM5B's role in breast cancer is still being actively investigated, emerging research suggests its involvement in breast cancer progression and metastasis \citep{di2023truncated}.



The RANBP1 gene, also known as Ran-binding protein 1, encodes a protein involved in the regulation of the Ran GTPase cycle. The Ran GTPase is essential for nucleocytoplasmic transport, which controls the movement of molecules between the nucleus and the cytoplasm. While RANBP1 itself is not directly associated with breast cancer, aberrant expression or dysregulation of proteins involved in the Ran GTPase cycle, including RANBP1, have been implicated in cancer, including breast cancer \citep{yuen2016ran}\citep{bamodu2016aberrant}. Some studies have suggested that RANBP1 may have tumor suppressor properties. Decreased expression of RANBP1 has been associated with poor prognosis and aggressive features in breast cancer patients. Loss or downregulation of RANBP1 expression may contribute to tumor development and progression.



%

\begin{figure}[ht]
  \centering
  \includegraphics[scale=0.67]{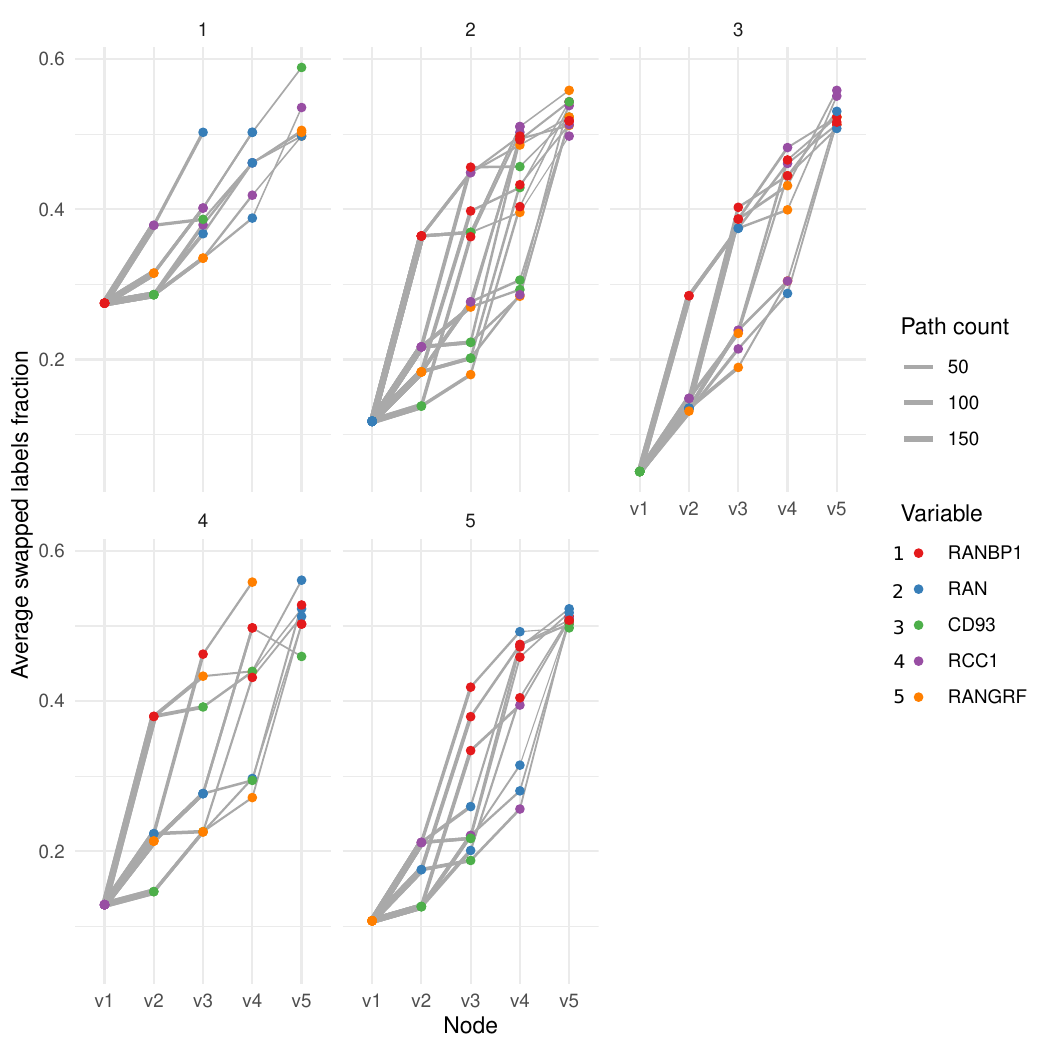}
  \caption{Breast cancer dataset.  Graphical summary of the counterfactual paths.}
   \label{fig:PPI2}
\end{figure}

In a further investigation, we retrieved the first-order neighbourhood (N) of the detected and aforementioned genes. The results were AUC(N(LAT))= 0.78, AUC(N(RANBP1))= 0.82, AUC(N(KDM5B))= 0.67. The neighborhood of the RANBP1 gene (N(RANBP1)) included four additional genes, namely RAN, CO93, RCC1, and RANGRF. For this subset, we repeated the generation of CPATH explanations, but this time without incorporating domain-knowledge. The number of paths was set to $1000$, and the length of paths was $k=5$. We obtained 864 counterfactual paths. The generated counterfactual paths can be obtained from Figure \ref{fig:PPI2}. The shortest counterfactual path goes through RANBB1, RCC1 and RAN, leading to an average of 50\% label swaps.

For the first-order neighborhood of RANBP1 we applied bayesian network learning based on the detected counterfactual paths. The RANBP1 strongly depends on the CD93 gene. When CD93 is part of a counterfactual path the probability is $0.74$ that RANBP1 also is included.


\section{Discussion}
Apart from the proposed technique, various other approaches can be used to obtain importance values based on sampling trajectories of finite-length random walks on the $\widetilde{G_{\mathcal{M}}}$ graph (with corresponding changes in model predictions). In particular, other approaches can use any sampled trajectory to compute importance scores without relying directly on the definition of counterfactual paths using the counterfactual policy $\Psi$. Therefore, in our initial experiments, we also tested solutions based on reinforcement learning (RL) methods. The main difference with the technique presented earlier is that in reinforcement learning-based approaches, successive sampled trajectories can be used to update the transition matrix (or directly the importance vector) iteration by iteration, after each episode of the learning process. 

When RL techniques are used to estimate feature importance, the agent learns a policy that maximizes the cumulative reward over time. The agent explores the graph by traversing different paths and observing the resulting changes in the model predictions. Technically, we can formulate the problem as a Markov decision process (MDP), which is a tuple $\langle S, A, P, R \rangle$, where
\begin{itemize}
    \item $S$ represents the state space corresponding to the set of vertices in the graph $G_{\mathcal{M}}$ (explanatory variables).
    \item $A$ is the action space, representing the available actions that an RL agent can take in each state. In this case, the actions correspond to traversing from one vertex to another in the graph.
    \item $P$ denotes the transition probabilities to other states, given the current state and the chosen action. In this particular environment, the state to which an action leads is deterministic (since the action is to select the appropriate edge of the graph).
    \item $R$ represents the reward function that provides feedback to the RL agent based on its actions. In the context of feature importance, the reward can be defined to reflect the changes in model predictions induced by traversing different paths.
\end{itemize}
The terminal state is then determined by a fixed maximum number of variables on the path or is associated with exceeding the threshold of the reward (significant change in the model predictions).

In our implementation, a solution using the model-free Q-learning algorithm is available. It estimates the matrix $Q_{p\times p}$ of state-action quality, from which the importance of variables can be extracted by aggregation, analogous to the main method. Other available techniques allow the direct estimation of the state value function (which translates into the importance of the variables) -- this can be done using the temporal difference learning TD(0) or the first-visit Monte Carlo algorithm. Although these methods give promising results, they suffer from sensitivity to hyperparameters. They require further research, investigation and validation, which is beyond the scope of this paper. 

Furthermore, our proposed approach was herein evaluated based on global explanations. A single test-instance, however, could be explained by first learning the transition matrix $\mathbf{T}$ on the training set. From this transition matrix candidate paths could be generated by a Markov process. The generated path features and their corresponding values could be exchanged with those in the training set to test for counterfactuals. 










\section{Conclusion}

We have developed a novel explainable AI method: counterfactual paths. Unlike classical feature importance methods, the generated explanations are efficiently visualized through graphs, which could help detect causal effects and important interactions between features. 

\section*{Code availability}

The herein presented method is implemented within the \texttt{cpath} package available for R and Python at \url{https://github.com/pievos101/cpath}.

\section*{Acknowledgements}

Parts of this work have been funded by the Austrian Science Fund (FWF), Project:  P-32554 “explainable Artificial Intelligence”
(Grantholder AH). Parts of this work have been funded by the Polish National Science Centre (NCN) grant 2019/34/E/ST6/00052 (Grantholder PB).

%
%
%
\bibliographystyle{spbasic}
\bibliography{references}
\clearpage

\appendix

\renewcommand\thefigure{ Appendix Figure \arabic{figure}}    
\setcounter{figure}{0}  

\renewcommand\thetable{ Appendix Figure \arabic{figure}}    
\setcounter{table}{0}  

\section{Appendix}
\subsection{Simulation Algorithms}

\begin{algorithm}[H]
\SetAlgoLined
 $n=100, p=4, D^{n\times p} \sim N(0,2), y^{n\times 1} \sim Bern(0.5)$\;
 $a=0, b=0$\;
 \For{$i=1$ \KwTo $n$ }{
    
  \eIf{$D[i, 1] \geq a$}{

    \If{$D[i, 2] \geq b$}{
        $y[i] = 1$\;
     }
   }{
    \If{$D[i, 2] \geq b$}{
        $y[i] = 0$\;
     }
   }
   }
 \caption{Simulation 1. Conditional dependency (1)}
\end{algorithm}

\begin{algorithm}[H]
\SetAlgoLined
 $n=100, p=4, D^{n\times p} \sim N(0,2), y^{n\times 1} \sim Bern(0.5)$\;
 $a=0, b=0$\;
 \For{$i=1$ \KwTo $n$ }{
    
  \If{$D[i, 1] \geq a$}{

    \eIf{$D[i, 2] \leq b$}{
        $y[i] = 1$\;
     }{
        $y[i] = 0$\;
     }
   }
}
 \caption{Simulation 2. Conditional dependency (2)}
\end{algorithm}

\begin{algorithm}[H]
\SetAlgoLined
 $n=100, p=4, D^{n\times p} \sim N(0,2), y^{n\times 1} \sim Bern(0.5)$\;
 $a=0$\;
 
 \For{$i=1$ \KwTo $n$ }{

    \If{$D[i, 1] \geq a \And D[i, 2] \geq a $}{
        $y[i] = 1$\;
     }
   
 }
 \caption{Simulation 3. Correlation}
\end{algorithm}

\begin{algorithm}[H]
\SetAlgoLined
 $n=100, p=4, D^{n\times p} \sim N(0,2), y^{n\times 1} \sim Bern(0.5)$\;
 $a=0$\;
 \For{$i=1$ \KwTo $n$ }{
    
    \If{$D[i, 1] \geq a \And D[i, 2] \geq a$}{
        $next$\;
     }
  
    \If{$D[i, 1] \geq a$}{
        $y[i] = 1$\;
     }
     
    \If{$D[i, 2] \geq a$}{
        $y[i] = 0$\;
     }
}
  
 \caption{Simulation 4. Conditional in-dependency}
\end{algorithm}

\newpage

\subsection{Supporting Figures}



\begin{figure}
     \centering
     \begin{subfigure}[b]{0.49\textwidth}
         \centering
         \includegraphics[width=\textwidth]{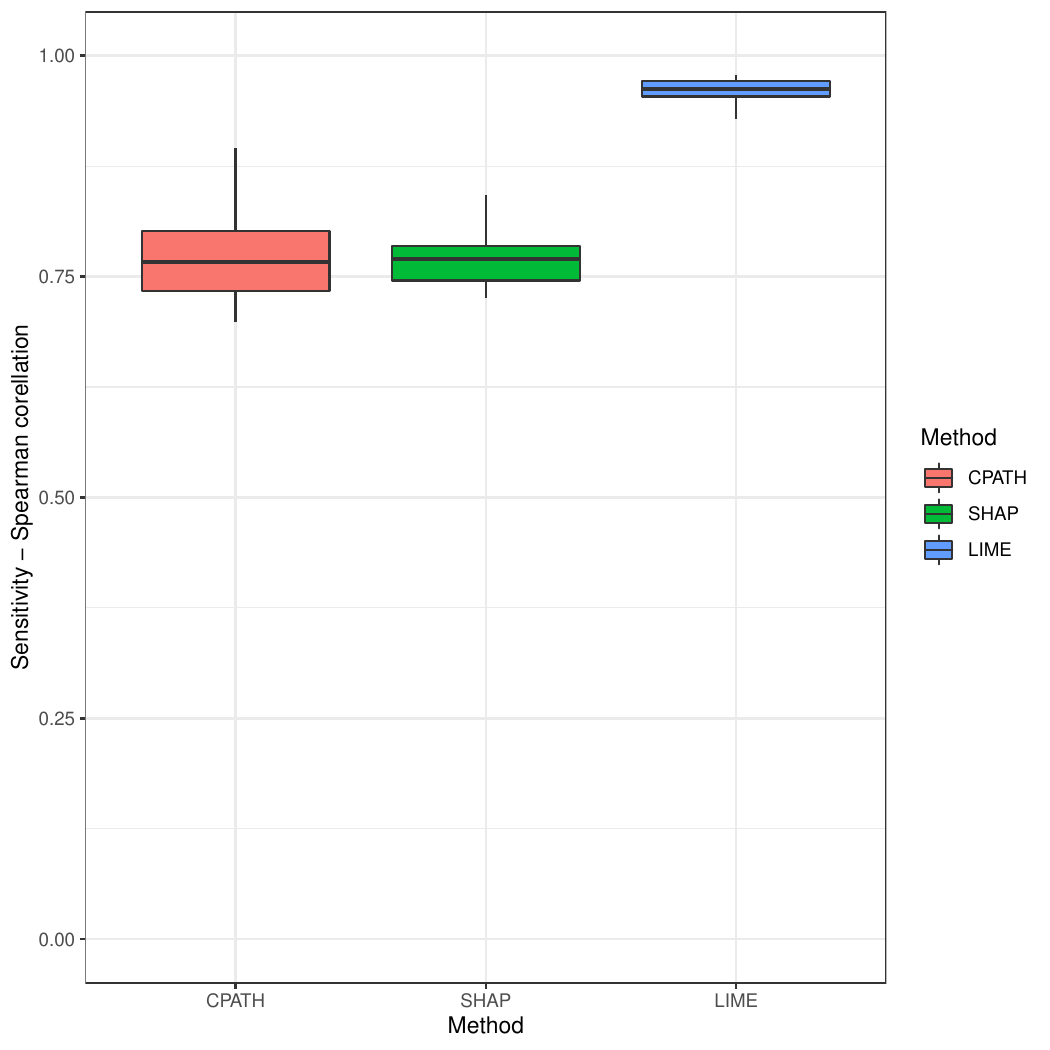}
         \caption{Ionosphere}
     \end{subfigure}
     \begin{subfigure}[b]{0.49\textwidth}
         \centering
         \includegraphics[width=\textwidth]{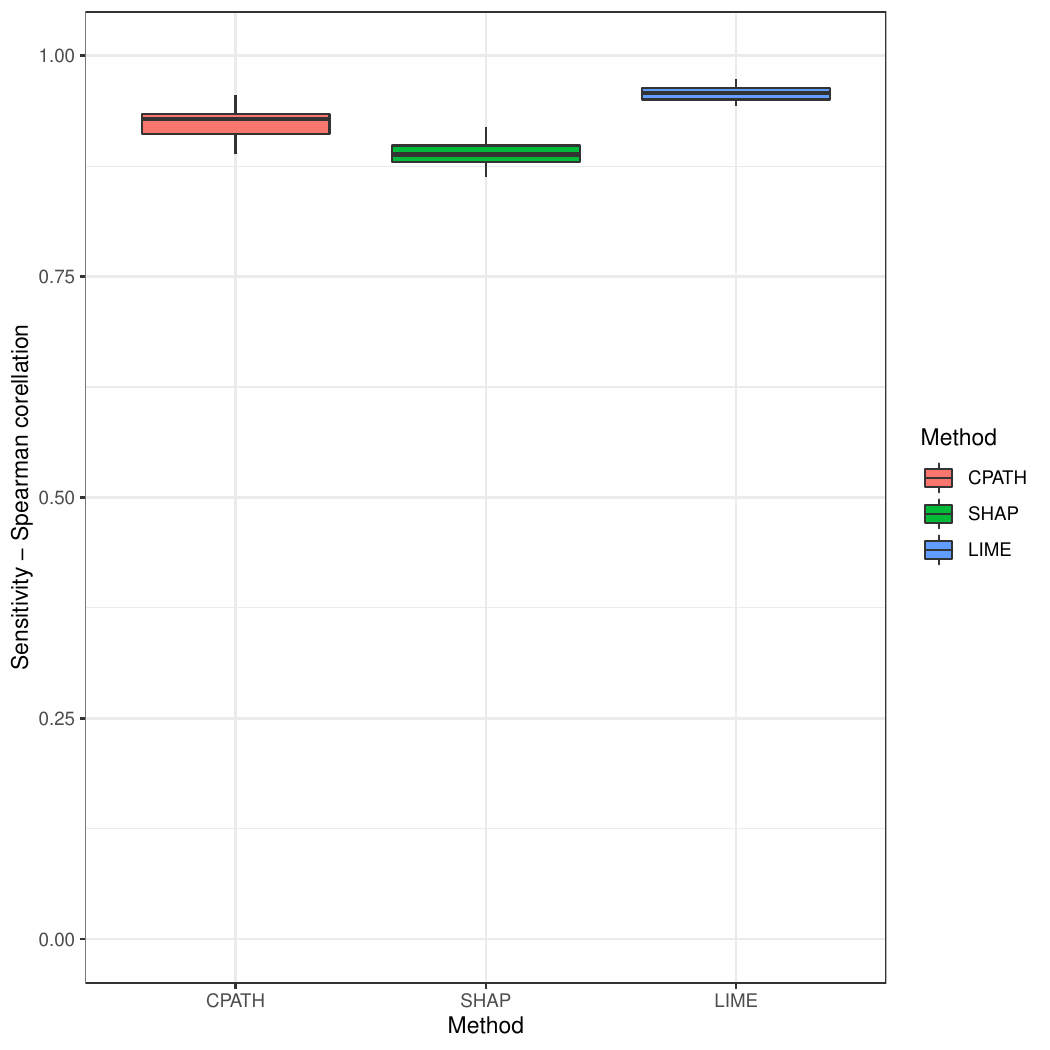}
         \caption{Breast cancer}
     \end{subfigure}
     \begin{subfigure}[b]{0.49\textwidth}
         \centering
         \includegraphics[width=\textwidth]{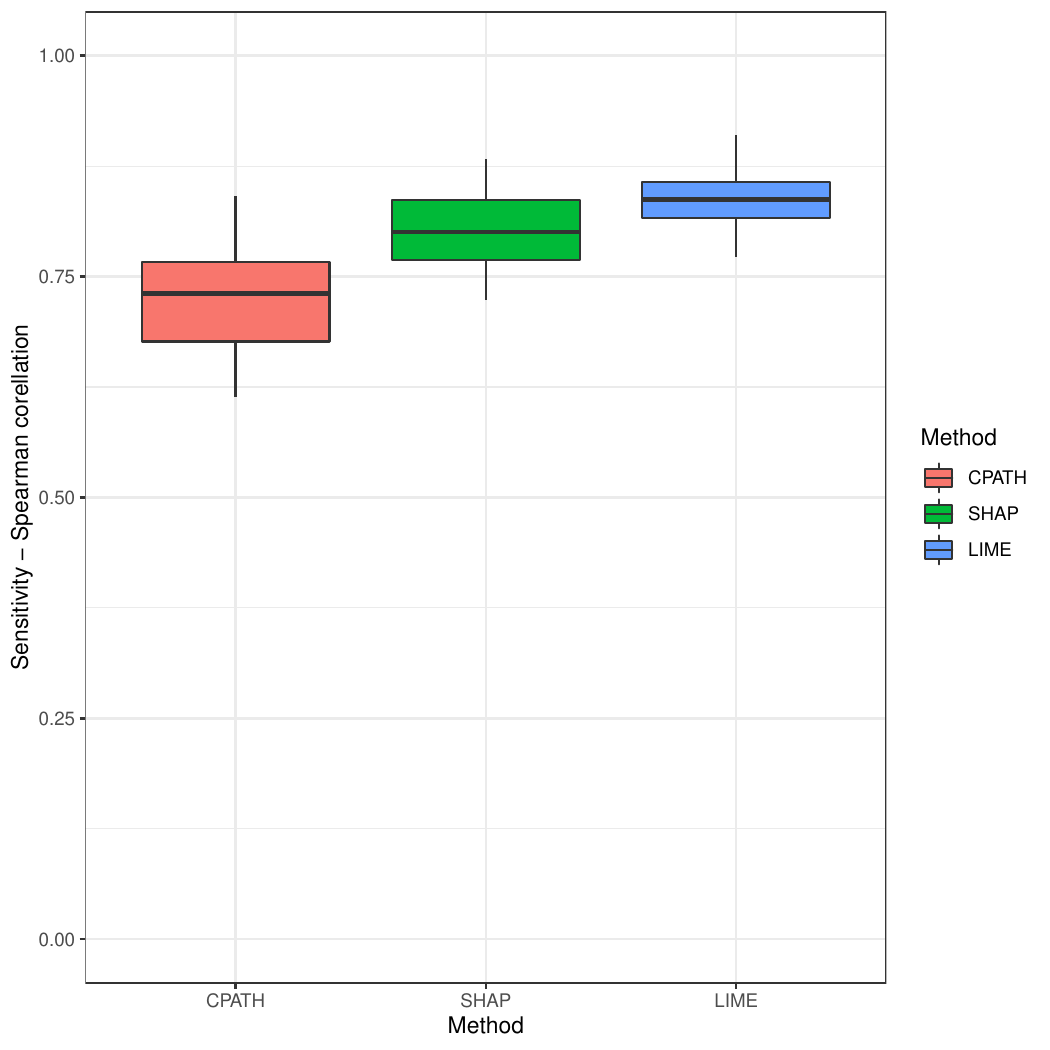}
         \caption{Diabetes}
     \end{subfigure}
     \begin{subfigure}[b]{0.49\textwidth}
         \centering
         \includegraphics[width=\textwidth]{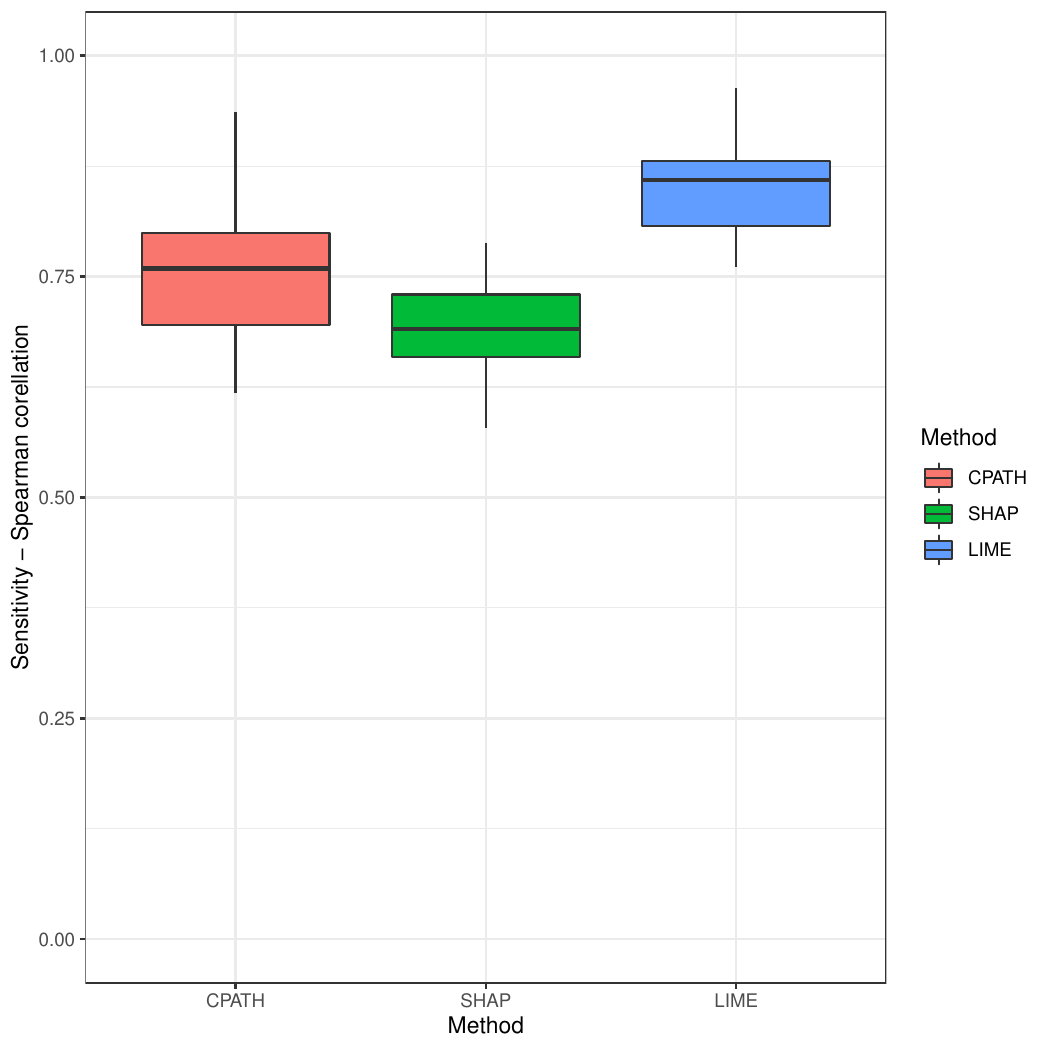}
         \caption{Iris}
     \end{subfigure}
        \caption{Sensitivity
       }
        \label{fig:sensitivity}
\end{figure}

\begin{figure}
     \centering
     \begin{subfigure}[b]{0.49\textwidth}
         \centering
         \includegraphics[width=\textwidth]{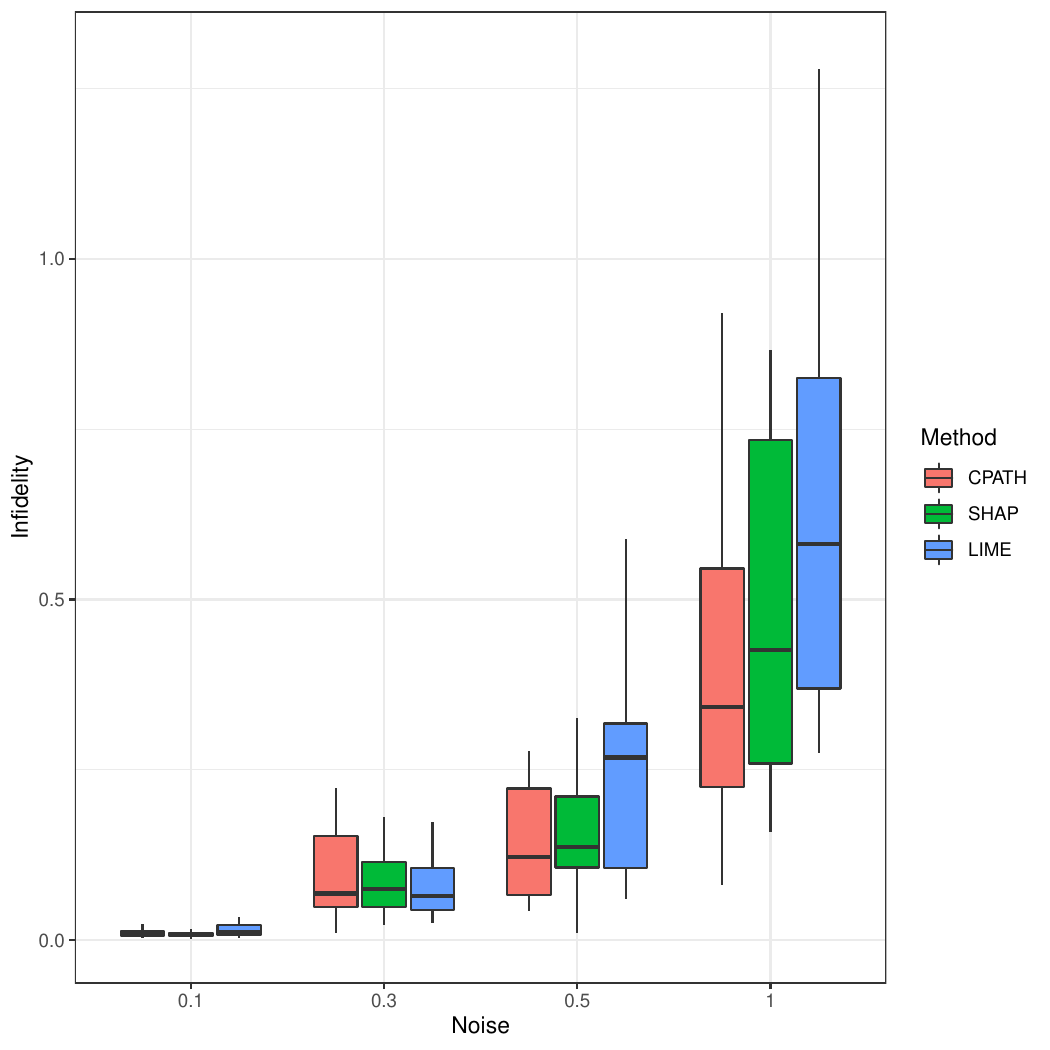}
         \caption{Ionosphere}
     \end{subfigure}
     \begin{subfigure}[b]{0.49\textwidth}
         \centering
         \includegraphics[width=\textwidth]{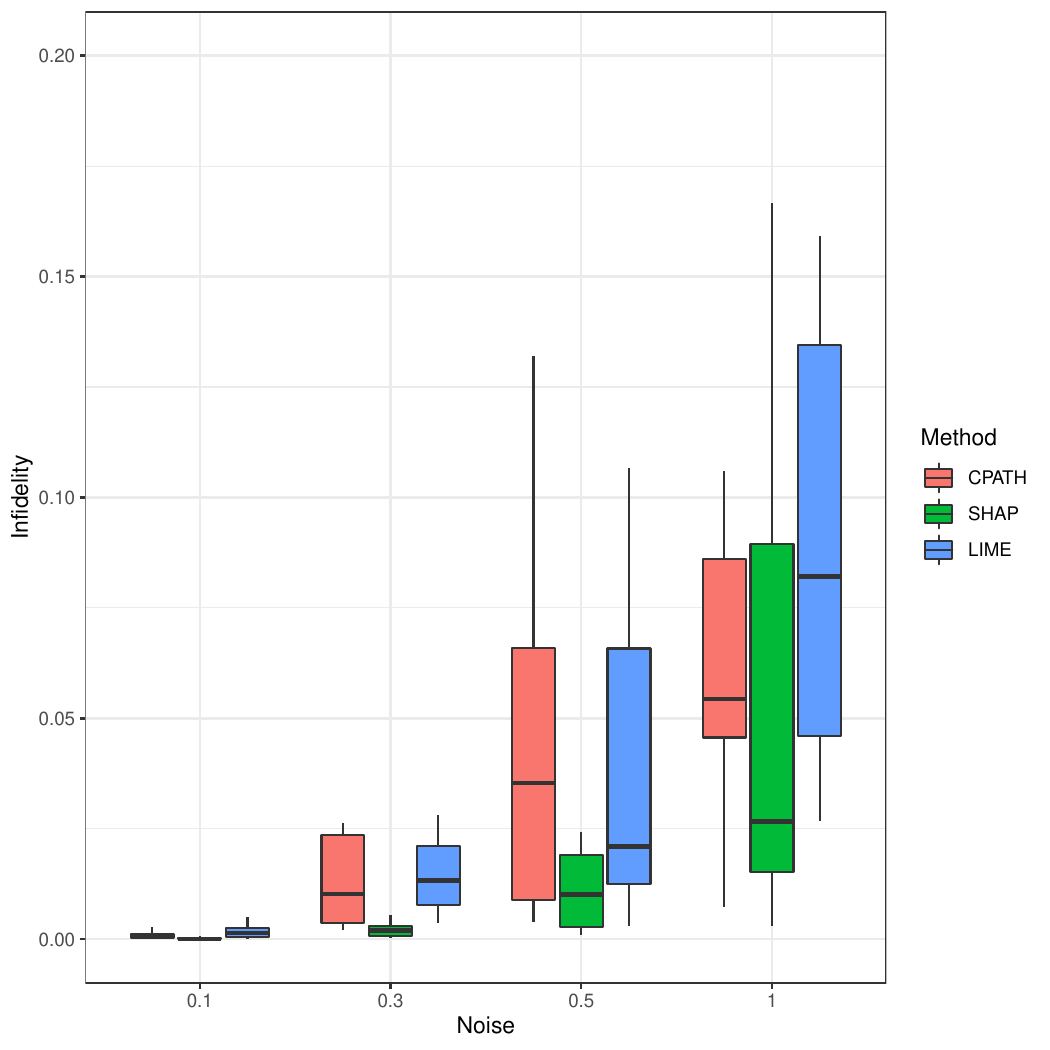}
         \caption{Breast cancer}
     \end{subfigure}
     \begin{subfigure}[b]{0.49\textwidth}
         \centering
         \includegraphics[width=\textwidth]{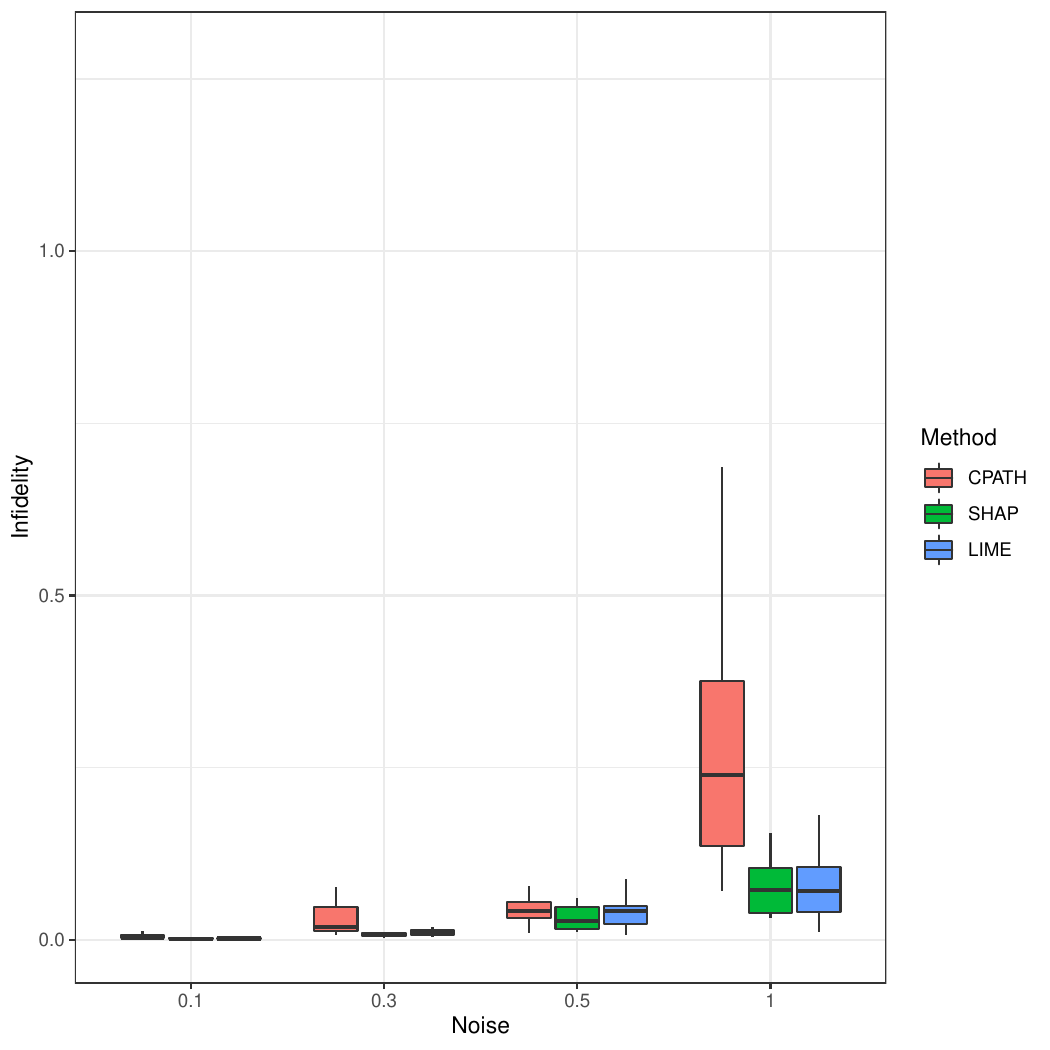}
         \caption{Diabetes}
     \end{subfigure}
     \begin{subfigure}[b]{0.49\textwidth}
         \centering
         \includegraphics[width=\textwidth]{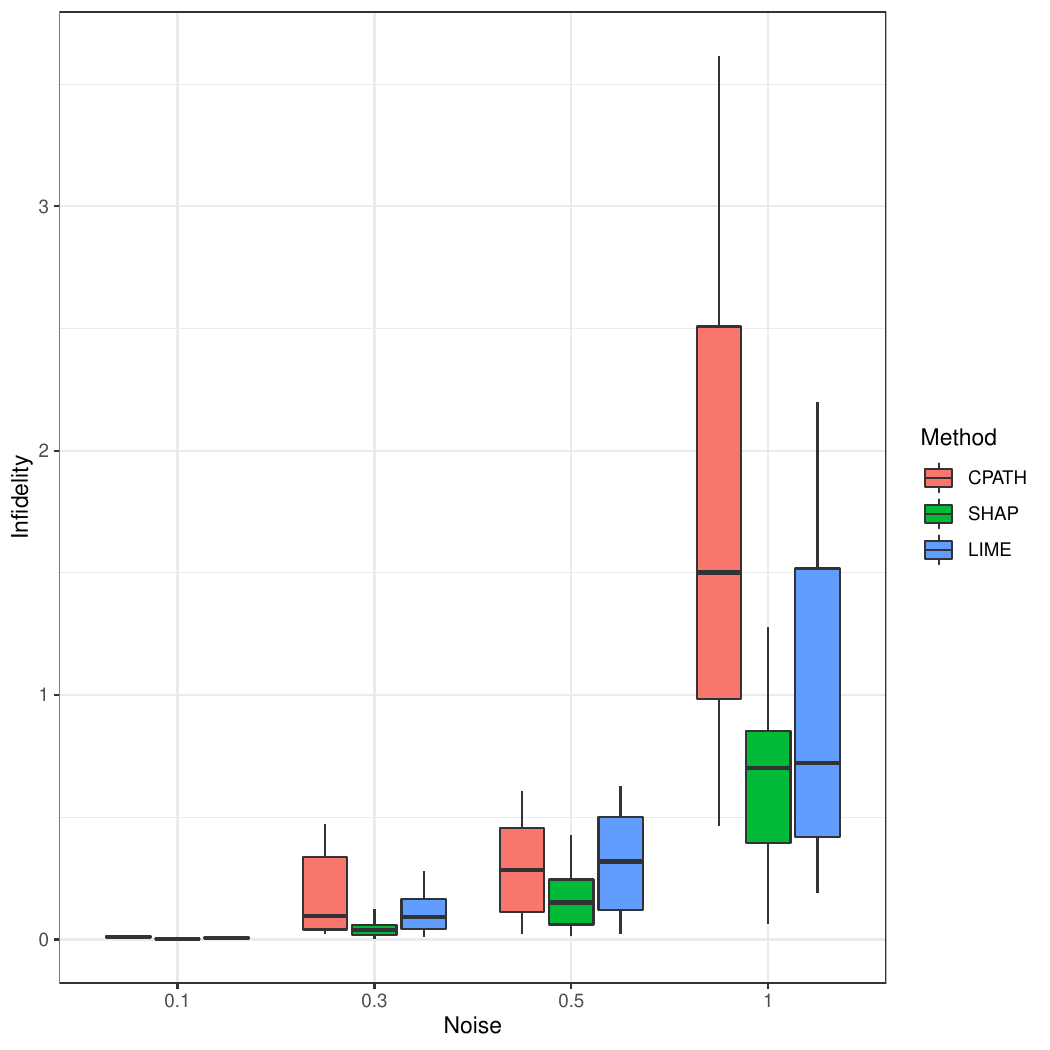}
         \caption{Iris}
     \end{subfigure}
        \caption{Infidelity 
       }
        \label{fig:fidel}
\end{figure}





\end{document}